\documentclass[sigplan,twocolumn]{acmart}
\usepackage{amsmath}
\usepackage{float}
\usepackage{url}
\usepackage{subcaption}

\usepackage{multirow}
\usepackage[most]{tcolorbox}
\usepackage{capt-of}
\usepackage{hyperref}
\usepackage{booktabs}
\usepackage{tabularx}
\usepackage{placeins}
\microtypecontext{spacing=nonfrench} 

\newcommand{\system}{\textsc{TinyLLM}}

\newcommand{\fakeparagraph}[1]{\vspace{.5mm}\noindent\textbf{#1.}}
\newcommand{\fakepar}[1]{\fakeparagraph{#1}}

\renewcommand\footnotetextcopyrightpermission[1]{}
\begin{document}

\title{\system: A Framework for Training and Deploying  Language Models at the Edge Computers}

\author [Kandala]{Savitha Viswanadh Kandala}
\orcid{0009-0009-0969-0953}
\affiliation{
  \institution{National University of Singapore}
  \country{}   
  }
\email{viswanadh@u.nus.edu}

\author [Medaranga] {Pramuka Medaranga}
\orcid{0009-0006-1512-4841}
\affiliation{
  \institution{National University of Singapore}
  \country{}   
  }
\email{pramukas@comp.nus.edu.sg}

\author[Varshney] {Ambuj Varshney}
\orcid{0000-0002-9282-4108}
\affiliation{
  \institution{National University of Singapore}
  \country{}   
  }
\email{ambujv@nus.edu.sg}

\renewcommand{\shortauthors}{Viswanadh and Ambuj}

\begin{abstract} 
Language models have gained significant interest due to their general-purpose capabilities, which appear to emerge as models are scaled to increasingly larger parameter sizes. However, these large models impose stringent requirements on computing systems, necessitating significant memory and processing requirements for inference. This makes performing inference on mobile and edge devices challenging, often requiring invocating remotely-hosted models via network calls. Remote inference, in turn, introduces issues like latency, unreliable network connectivity, and privacy concerns. To address these challenges, we explored the possibility of deviating from the trend of increasing model size. Instead, we hypothesize that much smaller models (~30-120M parameters) can outperform their larger counterparts for specific tasks by carefully curating the data used for pre-training and fine-tuning. We investigate this within the context of deploying edge-device models to support sensing applications. We trained several foundational models through a systematic study and found that small models can run locally on edge devices, achieving high token rates and accuracy. Based on these findings, we developed a framework\footnote{\url{https://tinyllm.org/}} that allows users to train foundational models tailored to their specific applications and deploy them at the edge.
\end{abstract}

\maketitle

\section{Introduction}
We have seen considerable interest in machine learning models based on transformer architecture~\cite{vaswani2023attention}. They are trained across modalities: Models trained on textual data have given rise to large language models~(LLMs)~\cite{devlin2018bert, llm_gpt2, llm_gpt3, llm_gpt4, llm_gemini}, which are now widely used for interaction with computers through chatbots. Similarly, models that are trained on images~\cite{rombach2022high} can now generate photorealistic visuals.  Models that can generate music are also trained on analog information like acoustic data~\cite{huang2018musictransformer}. Nonetheless, we find that a common thread across these models is the scaling of the training data size. This follows the observation that larger training datasets result in models that can provide more accurate responses while demonstrating general-purpose capabilities~\cite{emergent_abilities, scaling_laws, agi_gpt4}.

At a high level, a machine-learning model is defined by its parameters—weights that the model learns during training. The larger the model, the more parameters it has, and the more data it requires to effectively learn from training~\cite{scaling_laws, compute_optimal_LLM}. State-of-the-art (SoTA) models now reach hundreds of billions of parameters~\cite{llm_llama3, llm_gpt3}, posing significant challenges due to elevated computational demands. 

As parameter size grows, so do the memory and processing requirements for training and inference, limiting the feasibility of training and using these models on commodity computing systems. Typically, a model of tens to hundreds of billions of parameters requires clusters of expensive graphical processing units~(GPU) running for a prolonged period~\cite{carbon_emissions}, making this task highly challenging and infeasible for most people and organizations. For example, the Llama 3.1 70B variant required approximately 7 million GPU hours on Nvidia H100-80GB hardware, while the 405B variant required over 31 million GPU hours~\cite{llm_llama3}. Using 16,000 GPUs for pre-training, this translates to around 20 days of training for the 70B model and 78 days for the 405B model.

\begin{figure}[t!]
      \includegraphics[width=1\columnwidth]{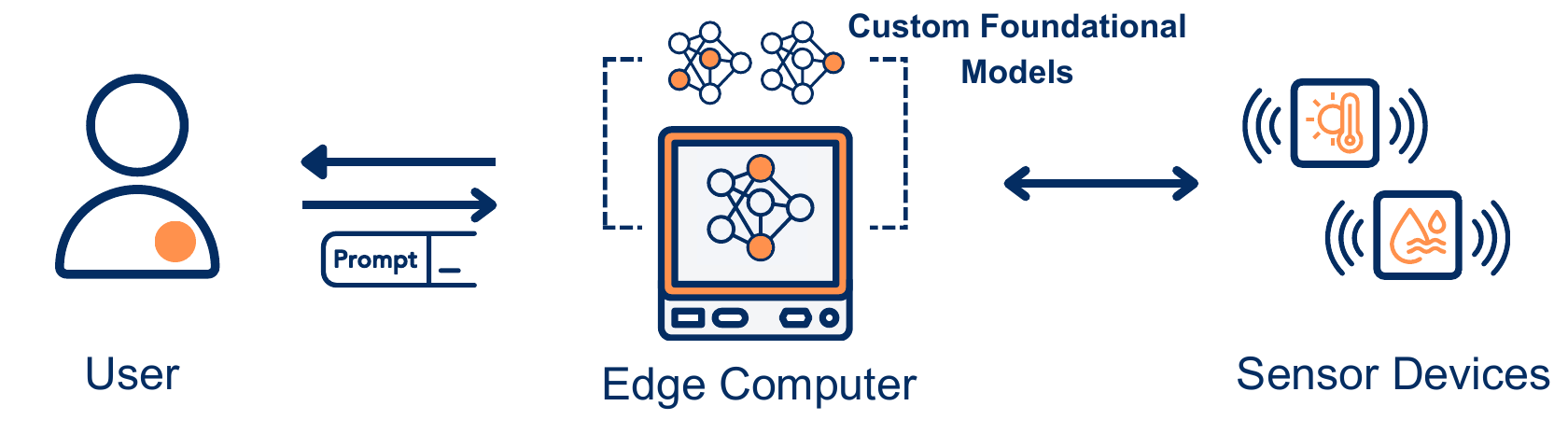}
  \vspace{-4mm}
  \caption{\emph{An embedded application often involves sensors that collect environmental data, which is then communicated to an edge device. \system\space provides a framework for training foundational models tailored for edge deployment, enabling these models to support a variety of tasks. This work explores training custom foundational models to enhance sensor data analysis. Our approach demonstrates a significantly smaller parameter-sized model than state-of-the-art language models, facilitating high-accuracy sensor data analysis while enabling rapid, local inference on even a constrained edge platform.}}
  \vspace{-4mm}
  \label{fig:overview}
\end{figure}

Beyond training, a larger parameter size model also negatively impacts the inference process. This is a step where the weights are loaded into memory and then used to answer queries through prompts provided by the user. As parameter size grows, so do the memory and processing requirements for inference, limiting the feasibility of using these models on commodity systems~\cite{llm_inferencing,memory_constraint, energy_llms}. For instance, loading a 70B parameter model in half-precision (FP16) would require at least 140GB of memory, exceeding the capacity of most GPUs. Today, even high-specification workstations struggle with memory bandwidth limitations, leading to the rise of alternative strategies to tackle model scaling~\cite{llm_llama3, squeeze_llm, smoothquant}. Consequently, the dominant mode of accessing models from mobile and edge devices has become function calls over a network to remotely hosted models. However, this approach introduces several challenges, including latency issues, unpredictable network conditions, and privacy concerns related to sharing sensitive information.


\begin{figure}[t!]
  \includegraphics[width=\linewidth]{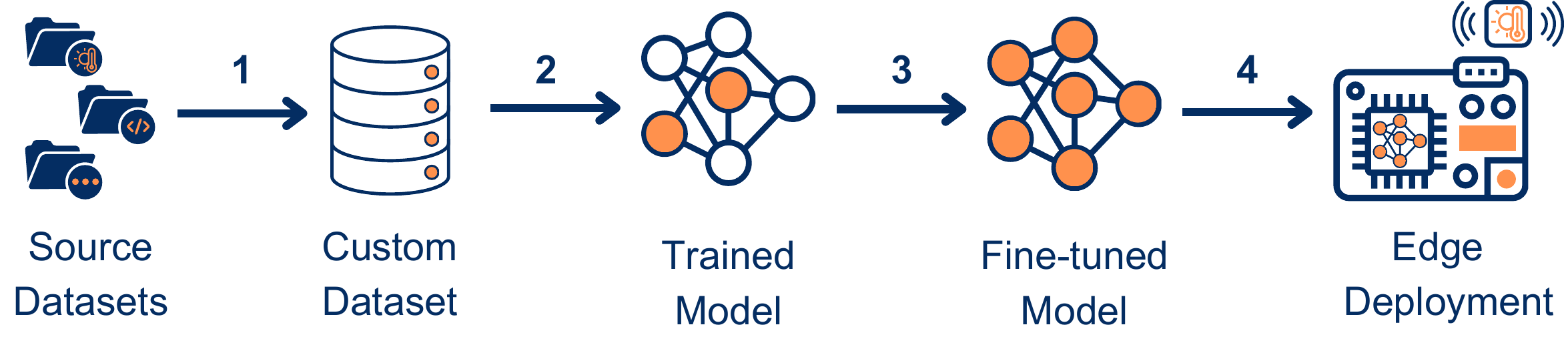}
  \vspace{-4mm}
  \caption{\emph{\system\space trains a custom foundational model for deployment at the edge device following a series of steps. It begins by appending a curated dataset with general conversational data. After pre-processing, the dataset is tokenized to pre-train a small model~(30-120M parameter). The pre-trained model undergoes fine-tuning with the custom dataset before deployment on the edge device to support embedded applications.}}
    \vspace{-4mm}
  \label{fig:training}
\end{figure}

A promising approach to tackle this challenge is explicitly trading off parameter size. A smaller parameter-sized model requires proportionally smaller memory and computing resources and can also perform inference faster, even on devices with constrained processing capabilities such as edge computers. \system\space framework builds on this approach.


\fakepar{\system\space Framework Overview}  We systematically study various trade-offs, models, and architectures and design a framework to pre-train foundational models from scratch. This framework is tailored to deploy such models at the edge. We prototype it for a challenging case related to embedded sensing, finding that much smaller models, with only tens of millions of parameters—orders of magnitude smaller than SoTA models—are sufficient for sensor data inference. We consolidate these insights into a framework called \system, meaning “more” in Swedish and “own” in Hindi. Pre-trained on carefully curated data, these smaller models offer significant benefits for embedded sensing applications. They can run locally on constrained edge platforms and perform rapid inference on modestly configured edge and mobile devices.

\fakepar{\system\space Design} \system\space simplifies the process for end-users to train custom foundational models for deployment at the edge. Users only need to provide a suitable training or fine-tuning dataset, and the framework manages the remaining steps to create a tailored foundational model. Performing this process involves following steps, as illustrated in Figure~\ref{fig:training}.

The first step involves preparing the dataset as the foundation for pre-training a model. Users can provide their dataset; however, even for smaller custom models, large amounts of data are typically required. It may be necessary to augment this data with relevant datasets related to language and conversation. \system\space facilitates this process by preparing the dataset for pre-training and offering a pre-curated collection that users can use to complement their datasets, ensuring an effective pre-training process.

The next step in the framework involves processing the data, which is crucial due to the diverse application scenarios requiring custom foundational models. For instance, in embedded sensing applications, even a simple sensor like an accelerometer may record motion across different axes. However, variations in resolution, sampling modes (analog vs. digital), and other intricacies can introduce inconsistencies. This step ensures data consistency by separating individual sensor readings by timestamp and organizing them into rows and columns. Additionally, it performs basic preprocessing tasks, such as removing unnecessary characters or spaces to fit the data within the model’s context window. Tokenization is the final step in the processing of the data.

Next, the framework involves training the foundational models. The framework adopts an architecture similar to existing models like GPT-2, which has proven effective for creating smaller models. Our results show that this approach achieves high accuracy for sensor data analysis. Additionally, the architecture allows flexibility in configuring parameter sizes as low as 30 million. While training these smaller foundational models still requires a GPU, the overall computing resources are minimal. For instance, we completed training on a single  Nvidia H100 in just a few hours.

After pre-training the model, we found that through extensive experiments, despite careful curation of the dataset, smaller models may still struggle to achieve high accuracy for specific applications. Therefore, fine-tuning becomes crucial to enhance their performance. The framework efficiently manages this step, requiring only a small set of examples for fine-tuning. Specifically, we employed the LoRa method for fine-tuning, significantly reducing the number of samples needed compared to traditional machine learning methods. For instance, in hand gesture sensing one of the use cases presented in the work. We only required 440 examples for the model to achieve high accuracy in detecting future events.


Once the custom model is trained and fine-tuned, it can be deployed on an edge device to support embedded sensing applications. While smaller models are generally expected to struggle with tasks involving mathematical operations and reasoning—key elements in many sensor data analysis tasks—we intentionally used these as a challenging test case for our system. Surprisingly, we found that smaller models can be highly effective for embedded sensing analysis and, in some cases, even outperform much larger models with significantly greater parameter sizes.

\fakepar{Summary of Results} The key results are:

\begin{itemize}
  \item We present a framework that supports two primary tasks: (1) training smaller models for edge deployment on user-defined datasets and (2) fine-tuning these models or off-the-shelf LLMs on domain-specific datasets. We demonstrate this by training five smaller models, ranging from 30M to 124M parameters, following the GPT-2 architecture. We also fine-tuned other models, such as Phi 2, Phi 3, and Llama 2, Llama 3.
  \item We compare the accuracy of smaller custom models trained through our framework, with fewer than 125M parameters, against larger models with billions of parameters across different IoT sensor datasets, including our collected and external datasets. Our results demonstrate that these smaller models perform comparably to larger ones while requiring significantly fewer GPU resources and less training time. 
  
  \item We investigate the suitability for deployment of smaller models on resource-constrained edge platforms and demonstrate that they lead to significantly faster inference or token generation rates.
\end{itemize}






\section{Background}
We provide the necessary background relevant to the design of \system. We also discuss and place related systems and developments related to the proposed system, \system.

\fakepar{Conditional probability view of models} Language modeling is framed as an unsupervised distribution estimation problem. Given a sequence of tokens $\mathbf{x} = [x_1, x_2, \ldots, x_n]$, the language model places a probability distribution $p(\mathbf{x})$ over the output token sequence. This probability can be decomposed into a product of conditional probabilities where each token depends on all the previous tokens:

\begin{equation}
    p(\mathbf{x}) = \prod_{i=1}^{n} p(x_i \mid x_1, x_2, \ldots, x_{i-1})
\end{equation}

This formulation allows the model to generate text by sampling from the distribution $p(\mathbf{x})$ and provides a basis for tractable estimation and sampling. The approach has been significantly improved by introducing models capable of computing these conditional probabilities effectively, such as the Transformer architecture \cite{vaswani2023attention}.

Causal language modeling proves effective for analyzing user prompts and generating text. Auto-regressive decoder models, such as GPT-2, are well-suited. These models, which \system\space trains for deployment on edge devices, excel in handling sequential data by predicting the next token based on previous ones, making them ideal for generating coherent text in response to user prompts.

\fakepar{Characterization of models} LLMs can be characterized along two primary axes: computational and memory requirements and performance. The parameter count—the number of weights a model contains—directly influences its computational and memory demands. Models can vary widely in size, ranging from hundreds of millions to billions of parameters. As parameter counts increase, so do the memory and processing resources required. Commercial vendors now offer models with parameter sizes reaching hundreds of billions, hosted on GPU-based clusters and accessible via web chat interfaces or API calls. Examples include ChatGPT~\cite{llm_ChatGPT}, Claude~\cite{llm_claude}, Gemini~\cite{llm_gemini}, and LLama~\cite{llm_llama2,llm_llama3}.

Performance-wise, two critical factors define LLMs: accuracy and response time. LLMs sometimes produce irrelevant or factually incorrect responses, referred to as hallucinations~\cite{hallucination}. Minimizing hallucination rates is a key goal, as higher rates negatively affect usability. Larger, cloud-based models tend to exhibit lower hallucination rates, with ongoing improvements targeting further reductions~\cite{hallucination_less}. Response time, measured by the token generation rate, refers to how many words or tokens the model generates per unit of time. A higher token rate translates to faster responses to user prompts.

\fakepar{Larger models are unsuitable for \system\space} Highly capable LLMs with tens to hundreds of billions of parameters present challenges, making them unsuitable for \system\space. \emph{First}, these models require expensive and complex infrastructure for inference, typically involving powerful computers with GPUs or custom ASICs. \emph{Second}, due to their high resource demands, they are hosted by third-party providers and accessed via API calls, leading to increased operational costs for end-users. \emph{Third}, sensor data, often containing private information, must be shared with these providers, posing significant privacy risks. \emph{Fourth}, fine-tuning these models is an expensive process requiring highly capable GPU-based machines. \emph{Lastly}, many of these models function as “black boxes,” raising concerns about using private data for training and fine-tuning without transparency.

\fakepar{Smaller models can run locally on edge computers} Smaller language models~\cite{llm_phi2,llm_phi3,llm_llama2,llm_llama3,llm_mistral,llm_gemma} trade parameter size for reduced computational and memory requirements, typically ranging from hundreds of millions to a few billion parameters—much smaller than their larger counterparts. This reduction leads to smaller model weights; for example, Microsoft Phi 2, with 2.7 billion parameters, has weights around 5.5 GB. A system running an LLM requires RAM at least as large as the model weight since the entire model must load into memory for inference. As a result, smaller models in half-precision (FP16) can run on edge-class commodity computers with 8–32 GB of RAM, such as Raspberry Pi~\cite{raspberrypi}, Lattepanda Sigma~\cite{lattepanda}, and Intel N100~\cite{intel_n100}.

There has been recent interest in smaller models, which inspire the design of \system.  Ma et al. \cite{1bitllm} introduces variants of LLM with ternary weights {-1, 0, 1}, effectively using just 1.58 bits per parameter instead of the usual 16-bit (FP16) or 32-bit floating-point (FP32), reducing the computational and memory bandwidths significantly while performing comparably with full precision transformer. Ruoss et al. \cite{chess_llm} shows a development of a 270M parameter LLM based on the transformer that achieves grandmaster-level chess rating, reaching the performance of best chess engines, showing that smaller and specialized models with careful training can perform at par or better than rule-based ML systems.  Srinivas et al. \cite{knowledge-boosting} introduces knowledge boosting, where a smaller model, usually deployed for real-time inferences on wearables, obtains delayed hints from a relatively larger model, often deployed on smartphones. This is shown in tasks involving speech separation and enhancement. There have also been works that have explored applying smaller models to mobile devices. Yuan et al. \cite{mobile-models} introduced the concept of a "mobile foundation model," which functions like firmware and can serve a wide range of  tasks on smartphones. This  model would be managed by the mobile OS and hardware and exposed as a system service to applications. 

\system\space allows for training at least an order of magnitude smaller language models consisting of only tens of millions of parameters. This enables faster inference even on constrained embedded platforms. From a training perspective, the end-user may easily train their custom models even with the modest computational resources. It also facilitates the deployment of models on a wide range of simpler embedded platforms with limited memory and RAM, including  SBCs, opening up the possibility of various embedded applications.

\fakepar{Challenges with remotely accessing larger models} To bring the capabilities of language models to edge devices, larger models can be accessed remotely. However, this approach introduces several challenges, particularly for embedded sensing applications, which are the focus of this work. \emph{First}, maintaining persistent network connectivity is often unfeasible, as many embedded sensing applications involve mobile devices, resulting in intermittent and unpredictable connections~\cite{persistent_network}. \emph{Second}, variable network latency and server constraints can lead to unpredictable response times, affecting the quality of service for time-sensitive tasks. \emph{Third}, remote model access incurs costs, with providers charging based on usage~\cite{openai_pricing}. \emph{Fourth}, embedded sensing applications frequently collect sensitive user data, raising privacy and security concerns when sharing information with third parties~\cite{inversion_attack}. \emph{Lastly}, while SoTA models offer powerful general-purpose capabilities, they can be excessive for specific embedded sensing tasks, which often do not require the full range of these models’ capabilities.

\fakepar{Our choice}  \system\space focuses on utilizing smaller models running locally on edge devices, driven by several key considerations. \emph{Firstly}, it enables executing the model closer to end devices, minimizing network dependencies and latency. Many end devices are situated in remote or hard-to-reach locations, where maintaining persistent network connectivity for the gateway device is challenging. \emph{Secondly}, many sensor applications involve collecting sensitive environmental data, and transmitting this data to third-party providers raises privacy concerns. Additionally, concerns persist regarding the use of such data to train LLMs. \emph{Thirdly}, local processing lowers operational costs, as third-party providers typically charge per-token usage fees. \emph{Finally}, as demonstrated in this work, smaller, task-specific models running locally can outperform generalized cloud-based models for specific applications, making them preferable over invoking more powerful but less specialized remote models.

\section{Design}

\begin{figure*}[t!]
\centering
  \includegraphics[width=0.85\textwidth]{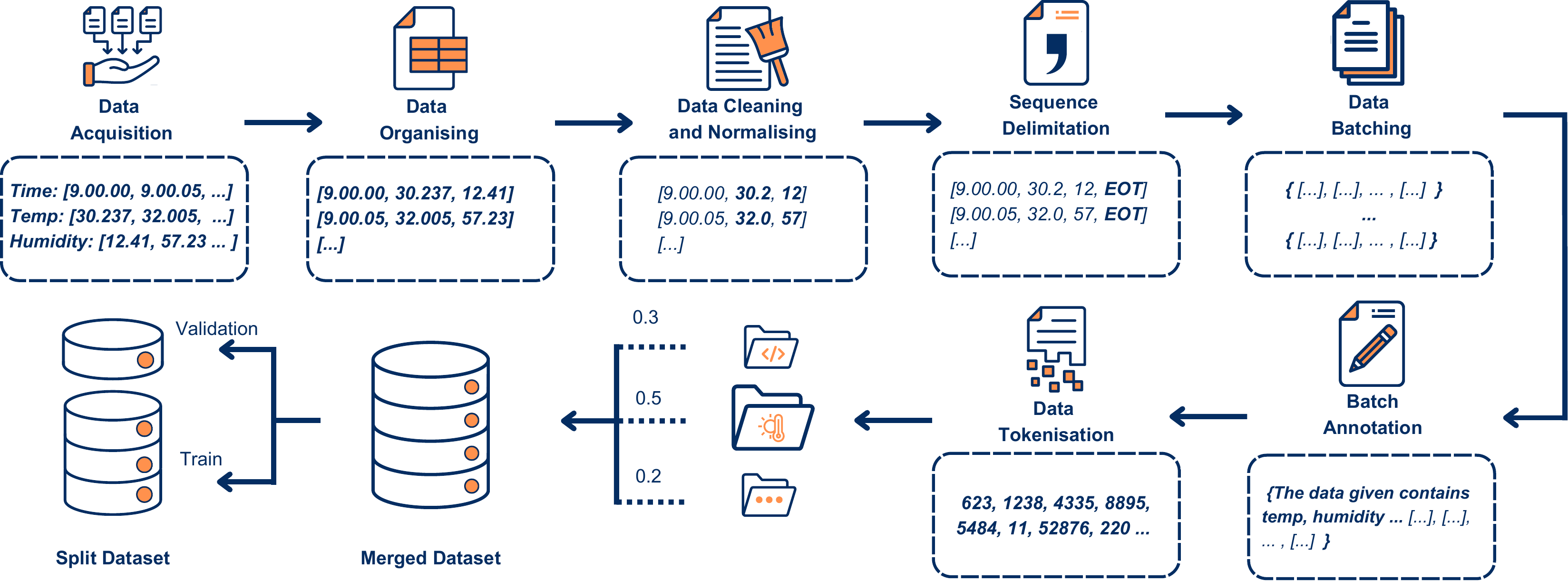}
  \vspace{-4mm}
  \caption{\emph{Processing the dataset is essential for effective pre-training. This step addresses the challenges posed by the dataset’s diversity, ensures alignment of the dataset with the model’s context window size limitations, and formats the data appropriately for its usage with the subsequent training process.}}
    \vspace{-4mm}
  \label{fig:dataset-prep}
\end{figure*}

\system\space  enables the training of custom foundational models for embedded sensing for deployment on edge devices.  It manages various process stages, including pre-processing diverse datasets to facilitate pre-training for a foundational model tailored for embedded sensing applications. \system\space also provides tools for fine-tuning models to align with the application scenario. The final step involves training and deploying the model on an edge device. We illustrate various steps in the \system\space in the Figure~\ref{fig:training}, and describe them next.

\subsection{Dataset preparation for pre-training of a model}

Training a foundational model requires a large corpus of data. When relying on third-party foundational models, users typically have very little control over the datasets used in the pre-training process, creating several challenges. \emph{First}, users are unsure if datasets relevant to their specific use case are included in the model’s pre-training, often leading them to opt for larger models with a higher likelihood of containing such information. This, however, increases parameter size and the computational requirements for inference. \emph{Second}, the opaque nature of these models raises legal and compliance concerns, as users cannot verify whether the model was trained on copyrighted or restricted content. \emph{Finally}, third-party models may also incur additional costs or impose licensing restrictions, further limiting their usability.

\system\space enables users to curate the dataset used for pre-training, giving them greater control over the information included in the model's training process. This approach reduces non-essential information being part of the pre-training dataset, resulting in the generation of highly specialized, smaller models tailored for the specific application scenario.

\subsection{Processing of the pre-training dataset}

After preparing the dataset for pre-training, the next step in the framework involves curating it for its efficient use during the training. This step is crucial since datasets often come in diverse formats and structures, which must be handled properly before use. Additionally, models have limited context windows, so the data must be structured to maximize the efficient use of this window. Finally, the dataset is tokenized to ensure compatibility with the training process. The processing step ensures data consistency for subsequent usage in the training process. We illustrate the various steps involved in the processing stage of the framework in  Figure~\ref{fig:dataset-prep}.

\fakepar{Transformation} The dataset structure often varies based on the application, with much of it consisting of numerical data. The first step focuses on transforming data into text since models are primarily trained using textual information. If the data contains timestamps, it may be grouped by timestamp into separate rows and columns. The transformation also involves data cleaning to optimize usage within the model’s limited context window. For example, 
GPT-2's 1024-token context window can limit longer prompts. We address this by normalizing readings to integers within specific ranges (0 to 100), reducing each reading to two characters. The framework adjusts character counts based on use cases and available context windows, keeping data compact and suitable for model training. The specific steps would vary by application and the model that is employed.


\fakepar{Tokenization} Tokenization prepares datasets for pre-training by converting text into numerical data that models can process. It breaks text into tokens, which are smaller units representing words, characters, or subwords. \system\space  uses the GPT-2 tokenizer to process the datasets. The tokenizer processes tokens organized into data shards by \system\space and marks row separations with end tokens. For unstructured datasets, we fill shards sequentially until the file is completed. For structured datasets with multiple columns, we merge columns into single prompts. Each prompt begins with dataset context, including information about units and data range, followed by data output and an end token. To optimize memory usage, we merge, tokenize, and store all prompts in fixed-size shards (200 MB). When tokens exceed shard size, we create new shards from the excess tokens.

\fakepar{Splitting and Mixing} The final step before pre-training involves mixing tokens from multiple datasets. Users can define the appropriate proportions for different types of data. For example, if the dataset includes textual conversations, the user can specify the proportion of tokens from such conversations. 
Using lazy loading, the framework loads the tokenized datasets in chunks, efficiently managing large data sizes. Tokens from different datasets are sampled based on user-defined ratios. A probability-based sampling method ensures that tokens are randomly selected from each dataset, maintaining the specified proportions. The selected tokens are then accumulated and saved into fixed-size shards, following the same process as in earlier steps. By default, \system\space splits the dataset into training and test sets in a 98:2 ratio.


\subsection{Training a custom foundational model}
Next, we describe steps to process a custom foundational model based on the processed dataset. We discuss the architecture of the model that we pre-train in this work.

\fakepar{Model architecture} We design the model architecture based on GPT-2~\cite{llm_gpt2}, a transformer-based, decoder-only language model that generates text by predicting the next word in a sequence. This architecture effectively supports training smaller parameter-sized models, which motivated our choice. Specifically, by adjusting the number of transformer blocks, the model offers variants with parameter counts between 30M and 124M. Additionally, \system\space ensures flexibility, allowing users to select and implement other architectures.

\begin{figure}[t!]
\centering
  \includegraphics[width=0.75\columnwidth]{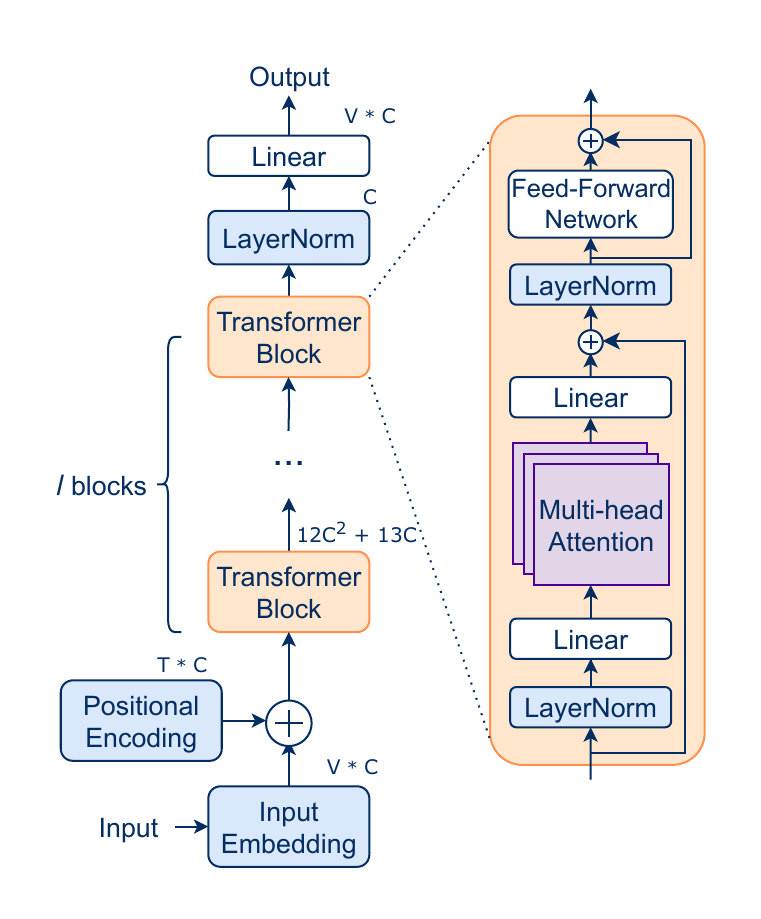}
  \vspace{-4mm}
  \caption{\emph{The high-level representation of the architecture for the model used in this work is based on the GPT-2. The model architecture consists of \textit{l} transformer blocks}}
    \vspace{-4mm}
  \label{fig:llm-arch}
\end{figure}

The architecture consists of an input embedding layer, positional encoding, multiple transformer blocks (each containing multi-head attention and feed-forward networks), and a final output layer as shown in the Figure~\ref{fig:llm-arch}.

\emph{Input Embedding:}\space Converts input tokens into dense vectors of fixed size $C$. These vectors represent the input tokens in a high-dimensional space where similar words have similar vectors. The input tokens are characterised by the vocabulary size ($V$), which is the number of unique tokens (words or subwords) the model can recognize and generate.

\emph{Positional Encoding:}\space This layer adds the positional information to the token embeddings so that the model can distinguish between the positions of tokens in a sequence. This is important since the transformer architecture itself does not inherently encode order information.

\emph{Transformer Block (Repeated $l$ Times):}\space Each transformer block consists of multiple sub-blocks which are as follows: 1) {Layer Normalization:} Normalizes the inputs to the block to stabilize and speed up the training process. 2) {Multi-Head Attention:} Applies self-attention to the inputs, allowing the model to simultaneously focus on different parts of the sequence. This is done in multiple "heads" in parallel, each learning different aspects of the sequence. 3) {Feed-Forward Network (FFN):} Consists of two linear transformations with a non-linearity (Gaussian Error Linear Unit (GeLU)) in between. It helps capture complex patterns by transforming the output of the attention mechanism. 4) {Residual Connections and Additional LayerNorm:} Adds the input of each sub-layer (Attention and FFN) to its output to form a residual connection. This  helps in stabilizing the learning process.

\fakepar{Training process} To train a model, the \system\space builds ontop of llm.c, that provides implementation of GPT-2.  It consists of  $l$ transformer blocks, with the total number of parameters $N$  summarized: The input embedding layer has $V \times C$ parameters, and the Positional Encoding layer has $T \times C$ parameters.
Each transformer block contains layer normalization, multi-head attention, and feed-forward network components. The total parameters per block are $l \times (12C^2+12C)$.
With $V = 50,257$, $T = 1,024$, $C = 64l$, we can estimate the total parameters of the model (in Million) approximately using the empirical expression:

\begin{equation}
N = 0.05l^3 + 3.2l
\end{equation}

By varying $l$, we can scale the model to different sizes, balancing model capacity with computational requirements.

\begin{table}[ht]
\centering
\resizebox{\columnwidth}{!}{%
\begin{tabular}{cccc}
\hline
Depth ($l$) & Hidden Size ($C$) & Parameters ($N$) & RAM Usage (MB) \\
\hline
6  & 384 & 30M  & 95.49   \\
8  & 512 & 51M  & 148.37 \\
10 & 640 & 82M  & 219.27 \\
11 & 704 & 102M & 262.89 \\
12 & 768 & 124M & 312.7  \\
\hline
\end{tabular}
}
\caption{Approximate parameter count and RAM usage of the GPT-2 model for given depth and hidden size values}
\label{tab:gpt2_parameters}
\end{table}

Embedded platforms are often limited in memory and processing capabilities. Therefore, we intentionally select a parameter size between 30M and 124M, as shown in Table~\ref{tab:gpt2_parameters}, for pre-training the foundational model. As demonstrated later, this choice enables the model to perform rapid inference even on constrained embedded platforms, such as SBCs which are commonly used as edge devices.

\fakepar{Parameters and resources} We default to a 12-layer model unless stated otherwise. The training parameters include a micro-batch size of 64 and a sequence length of 1024. The model runs over 10 billion tokens for one epoch, approximately 20,000 steps. The training process requires around 25GB of GPU memory. It was conducted on a single Nvidia H100 GPU and completed in roughly 9 hours. To ensure stable training, we incorporated techniques such as learning rate scheduling—with 700 warm-up steps followed by cosine decay—and gradient clipping with a maximum norm of 1.0. 


\begin{figure}[t!] 
\centering
\begin{tcolorbox}[
    enhanced,
    attach boxed title to top left={yshift=-3mm,yshifttext=-1mm, xshift=3mm},
    colback=white, colframe=black, boxrule=0.5mm, rounded corners,
    title=Prompt: Gesture Detection, fonttitle=\bfseries,
    boxed title style={colframe=black, colback=white, rounded corners, boxrule=0.5mm, fontupper=\bfseries\footnotesize, interior style={draw=none, fill=none}}
]
\textbf{\#\#\# Instruction:}\\
Sensor data values are provided in the following order: proximity, red, green, and blue light intensity values. Using these sensor values, determine the hand gesture performed. Give your answer only as Tap, Double, or Hold.\\
\\
\textbf{\#\#\# Input:}\\
\textit{Proximity: [2, 10, ..., 23] \\Red: [244, 243, ..., 20] \\Blue: [255, 255, ..., 255] \\Green: [200, 201, ..., 45]}\\
\\
\textbf{\#\#\# Response:}\\
Hold

\end{tcolorbox}
\vspace{-4mm}
\caption{We borrow a template from Alpaca for prompts and dataset entries required for fine-tuning a pre-trained model. Fine-tuning is an important step to ensure accurate responses to user queries for the specific  application scenario.}
\vspace{-4mm}
\label{alpaca_prompt}
\end{figure}

\begin{figure*}[ht]
\centering
  \includegraphics[width=\textwidth]{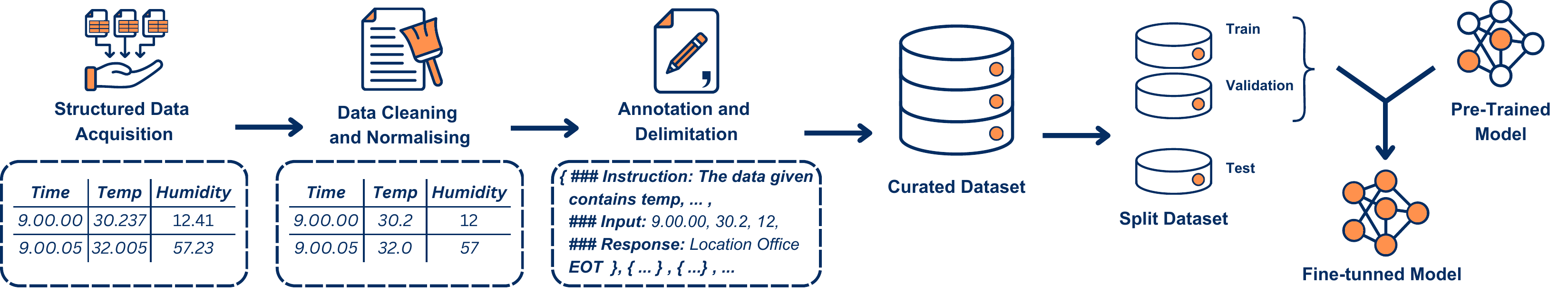}
  \vspace{-4mm}
  \caption{\emph{Pre-trained foundational models, despite careful dataset curation, often show lower accuracy. Fine-tuning these models with a small, curated dataset from the target application scenario significantly improves their accuracy. Our \system\space framework supports fine-tuning foundational models for deployment at the edge.}}
    \vspace{-4mm}
  \label{fig:fine-tuning}
\end{figure*}

\subsection{Finetuning of pre-trained custom model}
Even with curated datasets, we observed that the pre-trained model struggles to answer specific user queries~\cite{finetune}, even when similar examples exist within the pre-training data. Consequently, the model requires an additional fine-tuning step tailored to the specific application scenario. This step introduces new queries that the model has not encountered before and uses these examples to improve its ability to answer related queries. \system\space supports the fine-tuning of the custom-trained foundational model through several steps. 

The process begins with collecting relevant data for fine-tuning, requiring structured data formatted as input-output pairs. Each entry includes an input query and the expected output from the model based on the provided input. To optimize token usage and fit within the model’s limited context window, we apply min-max normalization to reduce the character count. Each dataset entry used in the process must be formatted according to predefined prompt templates. We adopt the Alpaca model template~\cite{alpaca}, which consists of three components: an instruction that describes the task, an input that provides additional context, and a response that completes the request, as illustrated in Figure~\ref{alpaca_prompt}. Each prompt must include relevant context within the prompt entry, thus ensuring a coherent input-output pair. The next step involves shuffling the dataset entries and splitting them into training, validation, and test sets. The split ratio depends on the size and nature of the data (e.g., 80\% training, 10\% validation, and 10\% test). This step ensures the model generalizes better by preventing overfitting to any specific order in the data. The final step involves the fine-tuning process.

Various techniques are available for fine-tuning, including PEFT (Parameter-Efficient Fine-Tuning). PEFT methods are popular because they allow users to train only a fraction of the model’s parameters, significantly reducing the memory footprint compared to full model fine-tuning. One such method is LoRA (Low-Rank Adaptation)~\cite{loraft}, which requires setting the adapter size and other parameters. As a result, we employ LoRA as part of the \system\space framework.

The key parameters for fine-tuning are the number of fine-tuning steps (or epochs), batch size, learning rate (which must be carefully tuned to prevent overfitting or underfitting), and dropout rates to control model regularization. Specifically for LoRA, the parameters include the adapter size, which determines how much the model adapts to new information; the rank of adaptation, which specifies the layers of the model affected by fine-tuning; and the scaling factor, which controls the contribution of various parameters.

\fakepar{Post-Fine-Tuning Evaluation:} Once fine-tuning is complete, the model generates domain-specific responses by passing queries using the  template employed during training.

\subsection{Implementation}
Embedded sensing applications often do not produce sufficient data to train a language model independently. General information must also be incorporated to enhance interaction through natural language prompts with such specialized models. We curated a base dataset of over 9 billion tokens from publicly available sources, which can be combined with user-provided datasets for pre-training the foundational model. This addresses scenarios where the available data from the user is limited. Specifically, we utilized the Fineweb dataset~\cite{fineweb_dataset_hf}, selecting 9 billion tokens from a collection of over 15 trillion tokens compiled from CommonCrawl dumps since 2013. Additionally, we incorporated the SHL dataset~\cite{shl}, which contains annotated data collected via smartphone sensors (e.g., accelerometers, gyroscopes, magnetometers, barometers, GPS) for human activity recognition across activities like walking, running, sitting, and driving. To further enhance the dataset, we included the ExtraSensory dataset~\cite{extra_sensory}, which comprises multi-sensor data from 60 participants over several days, totaling 300,000 minutes of activity and environmental context (e.g., walking, sitting, outdoors, at home) sampled every minute.

The mixed dataset integrates Fineweb and sensor datasets in user-defined proportions, totaling 9 billion tokens. We split the dataset into training and validation sets using a 98:2 ratio. For the SHL dataset, we merged sensor values corresponding to a given timestamp with descriptive prompts followed by the associated human action. We maintained consistency by splitting the datasets into 60-65\% for training, 20-30\% for testing, and 10-15\% for validation.

The Hugging Face Transformers library\footnote{\url{https://huggingface.co/docs/transformers/en/index}} was employed for fine-tuning. We used LoRA adapters with ranks varying in powers of 2 (from 16 to 256) and dropout probabilities between 0.1 and 0.3, with gradient accumulation over six steps. Learning rates ranged from $4e^{-4}$ to $6e^{-4}$, and the training steps varied from 100 to 300, using the AdamW optimizer~\cite{adamw} for fine-tuning. Gradient checkpointing was enabled, with evaluations performed at each logging step. The best model, determined by evaluation loss, was saved.

After fine-tuning, the base model was merged with the LoRA layers and converted to GGUF format, ensuring compatibility with the llama.cpp library~\cite{llama.cpp}. This conversion allows efficient inferencing on various embedded edge platforms through a C++ wrapper.
\section{Evaluation}
\begin{figure}[htb]
\centering
  \includegraphics[width=\columnwidth]{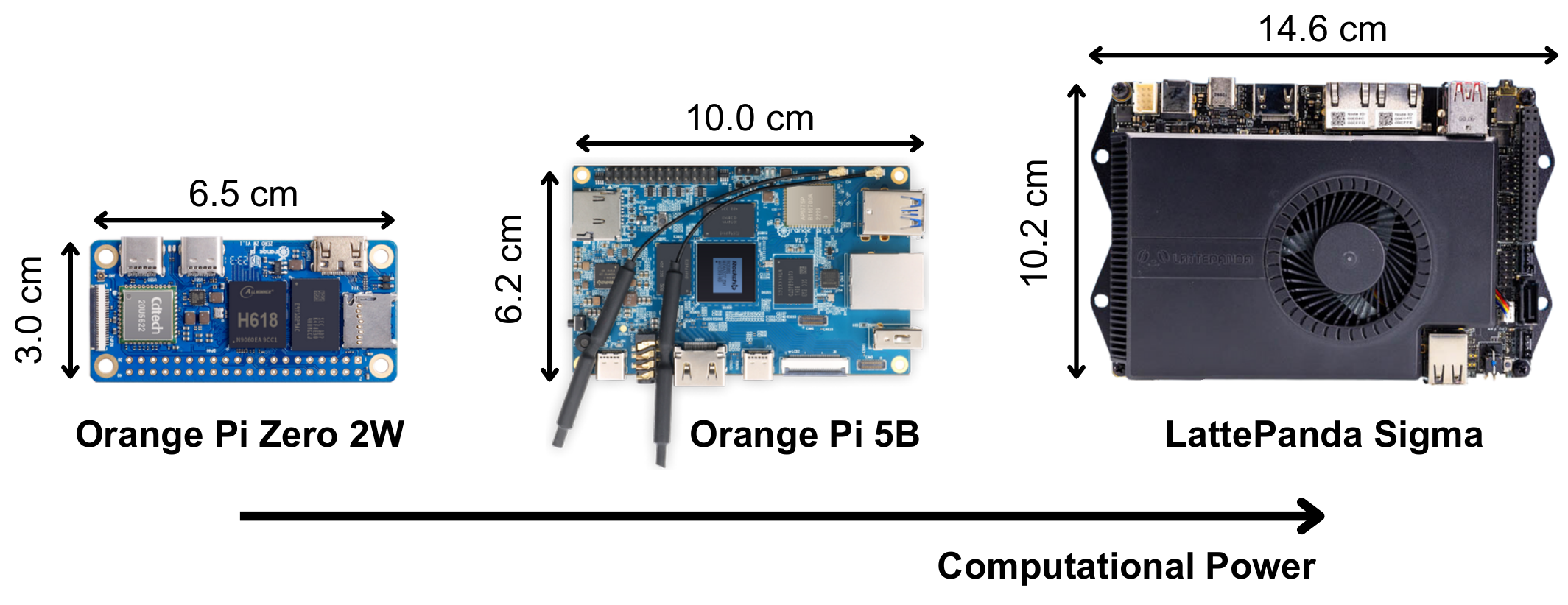}
  \vspace{-4mm}
  \caption{\emph{Embedded platforms for running \system\space-trained models vary in processing and memory capabilities, ranging from a few hundred megabytes to several gigabytes of RAM.}}
    \vspace{-4mm}
  \label{fig:single-board-computers}
\end{figure}

We evaluate the custom foundational models trained using \system across resource-constrained edge platforms. Specifically, we utilize single-board computers with diverse computational capabilities, as illustrated in Figure~\ref{fig:single-board-computers}. In these experiments, we focus on the impact of the pre-training data on the model's accuracy for embedded sensing applications. We compare the performance of our custom model against state-of-the-art language models, focusing on metrics such as token generation rate, task completion time, and accuracy.  The key highlights of some of the results are as follows. 

\begin{itemize}
\item Custom models exhibit improved performance after careful fine-tuning and incorporating domain-specific information into the pre-training dataset.
\item Custom models achieve better token generation rates and task completion times than commodity models
\item Smaller models can be deployed on resource-constrained edge platforms,  addressing memory limitations that hinder the deployment of commodity models.
\end{itemize}

\begin{table}[h]
\centering
\resizebox{\columnwidth}{!}{%
\begin{tabular}{ccccc}
\hline
\begin{tabular}[c]{@{}c@{}}Dataset\\ Name\end{tabular} & Source & \# Readings & \# Datastreams & \begin{tabular}[c]{@{}c@{}}\# Output\\ Labels\end{tabular} \\ \hline
\vspace{5pt}
\begin{tabular}[c]{@{}c@{}}Gesture\\ Detection\\  \end{tabular} & In-house & 630 & 4 & 3 \\
\vspace{5pt}
\begin{tabular}[c]{@{}c@{}}Localisation\\ \end{tabular} & In-house & 350 & 8 & 3 \\
\vspace{5pt}
\begin{tabular}[c]{@{}c@{}}Swimming Style\\ Detection  \end{tabular} & Brunner et al.~\cite{swimming_dataset} & 3,730 & 3 & 5 \\  \hline
\end{tabular}
}
\caption{Datasets used for evaluation. The datasets are split into 70\% for training, 10\% for validation, and 20\% for testing.}
\label{tab:datasets}
\vspace{-4mm}
\end{table}

\begin{figure}[htb]
\centering
\begin{subfigure}{1\columnwidth}
    \centering
    \includegraphics[width=0.7\columnwidth]{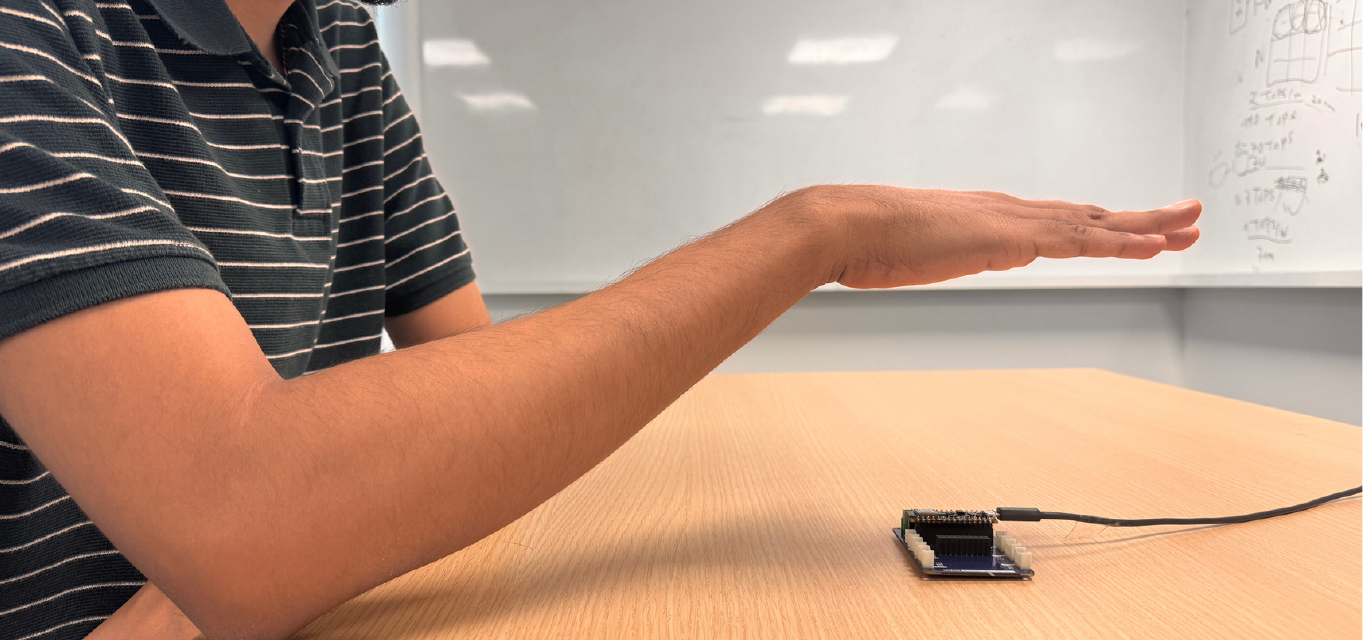}
    \caption{}
\end{subfigure}%
\vfill
\begin{subfigure}{1\columnwidth}
    \centering
    \includegraphics[width=0.85\columnwidth]{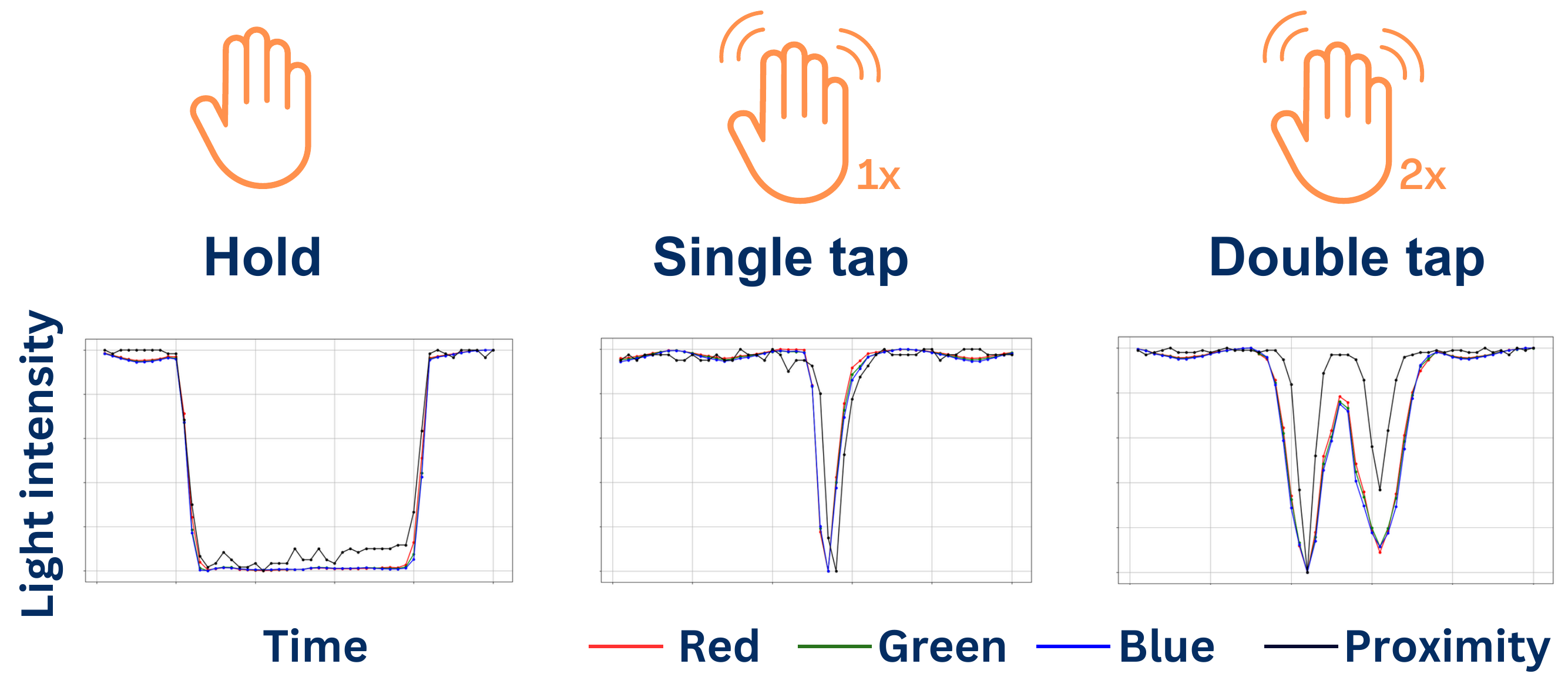}
    \caption{}
\end{subfigure}
\vspace{-4mm}
\caption{\emph{ (a) A user performing a hand gesture, and (b) observed light intensity values for different hand gestures.}}
\label{fig:gesture_dataset}
\end{figure}

\begin{figure*}[tb]
\centering
\begin{subfigure}{1\columnwidth}
    \centering
    \includegraphics[width=\columnwidth]{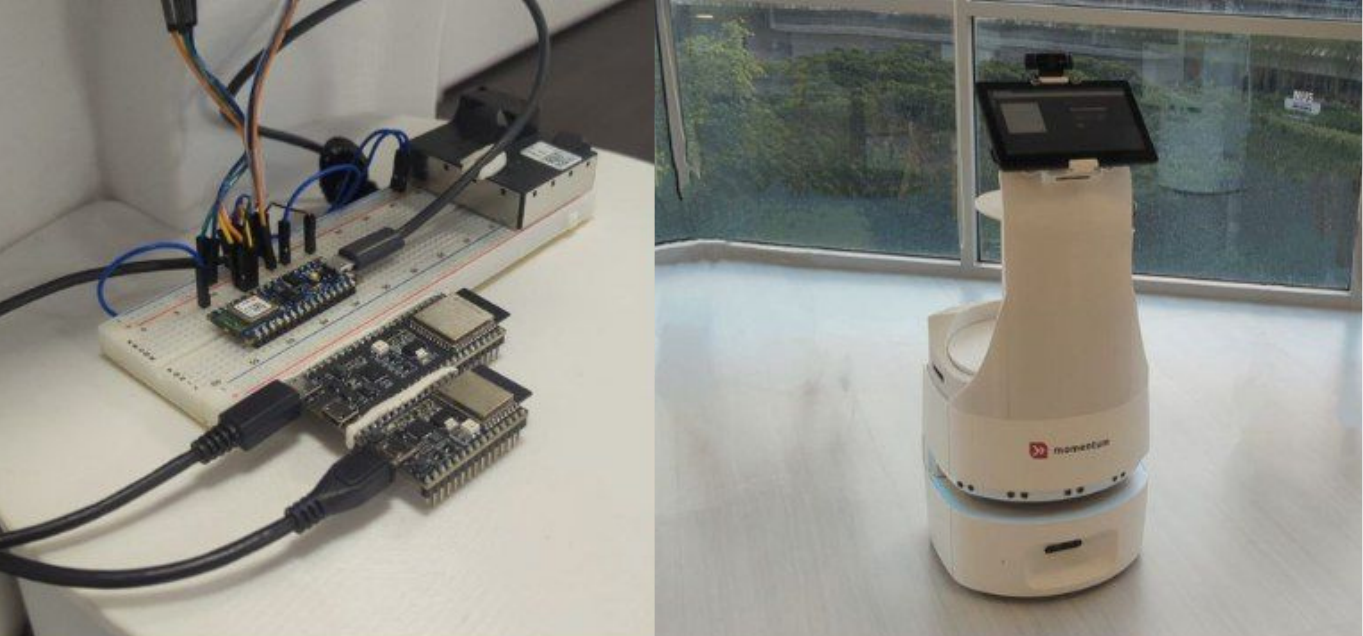}
    \caption{}
\end{subfigure}
\hfill
\begin{subfigure}{1\columnwidth}
    \centering
    \includegraphics[width=\columnwidth, height=4cm]{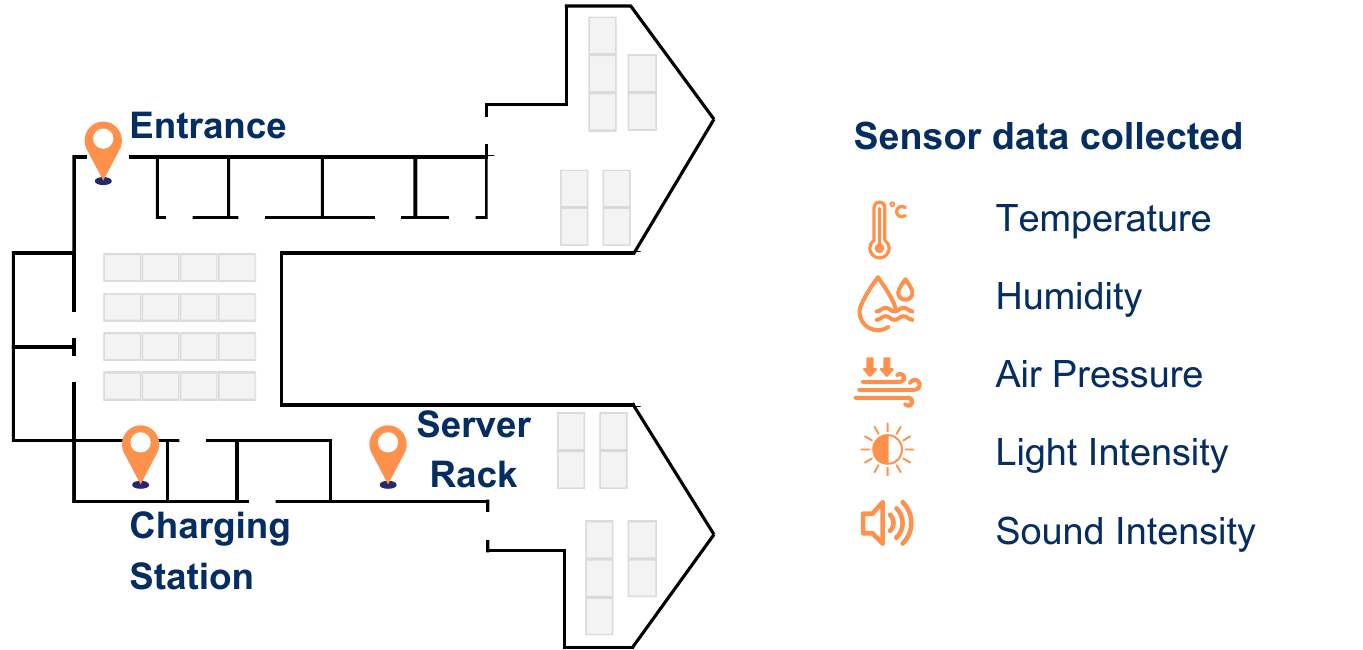}
  \caption{}
\end{subfigure}%
\vspace{-4mm}
\caption{\emph{The localisation dataset was created based on data collected from sensors deployed on a moving robot in an indoor workspace. (a) shows sensors deployed on an indoor robot for sensor-based location detection, and (b)The physical location and sensor data collected.}}
\label{fig:localisation_dataset}
\end{figure*}

\fakepar{Dataset} Since we target embedded sensing applications as a case study in this work, we collect relevant datasets to support such applications. Specifically, we utilize three datasets, summarized in Table~\ref{tab:datasets}. Two datasets—gesture detection and localization—were collected specifically for this work, while the swimming style detection dataset was sourced from~\cite{swimming_dataset} and is publicly available. It is worth noting that our framework can easily incorporate any additional datasets.

The custom dataset collected for this work includes 630 instances of hand gesture data captured using the APDS9960 light sensor at a sampling rate of 12.6 Hz, with each gesture instance lasting 4 seconds. This data was recorded from seven participants performing gestures under three distinct light levels: low (100–200 lux), medium (600–750 lux), and high (1500–1600 lux), as well as across two distance ranges: close (2–4 cm) and far (8–10 cm). The gestures included \textit{“Single Tap”}, \textit{“Double Tap”}, and \textit{“Hold”} (as illustrated in Figure~\ref{fig:gesture_dataset}), with an equal number of samples for each gesture class.

The second dataset captures environmental parameters characterizing various locations within a workspace (illustrated in Figure~\ref{fig:localisation_dataset}). The recorded parameters include temperature (°C), humidity (\%), air pressure (hPa), light intensity (RGB channels), and sound intensity. These measurements were taken at three distinct indoor locations: the entrance, the charging station, and the server rack. The dataset contains 350 instances collected using sensors mounted on a moving robot, with readings taken hourly between 11:00 AM and 5:00 PM over two consecutive days at the respective locations. The dataset comprises 120 instances each for the \textit{“Charging Station”} and \textit{“Entrance”} locations, and 110 instances for the \textit{“Server Rack”}.

Finally, the externally sourced dataset, the Swimming Style Detection Dataset~\cite{swimming_dataset}, consists of 17 hours of sensor data collected from 40 swimmers of varying skill levels. The data, obtained using the Nixon The Mission\footnote{\url{https://www.nixon.com/ch/en/smart}} smartwatch, includes measurements from an accelerometer, gyroscope, magnetometer, barometer, and ambient light sensor, sampled at a frequency of 30 Hz. We utilized only the accelerometer data for evaluations, comprising three streams representing movement along the $X$, $Y$, and $Z$ axes. These streams were segmented into chunks containing 100 readings per stream (approximately 3 seconds) to create input samples. Each chunk was annotated with one of five action labels, resulting in the following class distribution: \textit{“Freestyle”} (1,504 samples), \textit{“Breaststroke”} (285 samples), \textit{“Backstroke”} (475 samples), \textit{“Butterfly”} (222 samples), and \textit{“Transition”} (1,244 samples).

\fakepar{Pre-training and Fine-tuning Using Custom Models}
We provide a brief overview of the specific steps undertaken with these datasets for pre-training and fine-tuning model development within the \system framework. For pre-training, we first created a sensor dataset by processing data from the Extrasensory~\cite{extra_sensory} and SHL~\cite{shl} datasets, as illustrated in Figure~\ref{fig:dataset-prep}. This sensor dataset was combined with the Fineweb dataset, using a 40:60 split (unless explicitly stated otherwise) to pre-train the custom models for evaluation.

For fine-tuning, we followed the steps outlined in Figure~\ref{fig:fine-tuning}. Due to space constraints, we will only detail this process for the gesture dataset. The gesture dataset consists of four time-series data streams: proximity and light intensity (red, blue, and green channels). For each sample, sensor readings were concatenated into a text string formatted as follows:
\textit{Proximity: [...] \textbackslash n Red: [...] \textbackslash n Blue: [...] \textbackslash n Green: [...]}''.   The corresponding output labels included one of three gestures: \textit{Tap}’’, \textit{Double Tap}'', and \textit{Hold}’’. The instruction provided was:
``\textit{Sensor data values are provided in the following order: proximity, red, green, and blue light intensity values. Using these sensor values, determine the hand gesture performed. Give your answer only as Tap, Double Tap, or Hold.}’’

\begin{figure*}[t]
    \centering
    \begin{subfigure}[b]{0.42\textwidth}
        \centering
  \includegraphics[width=\textwidth]{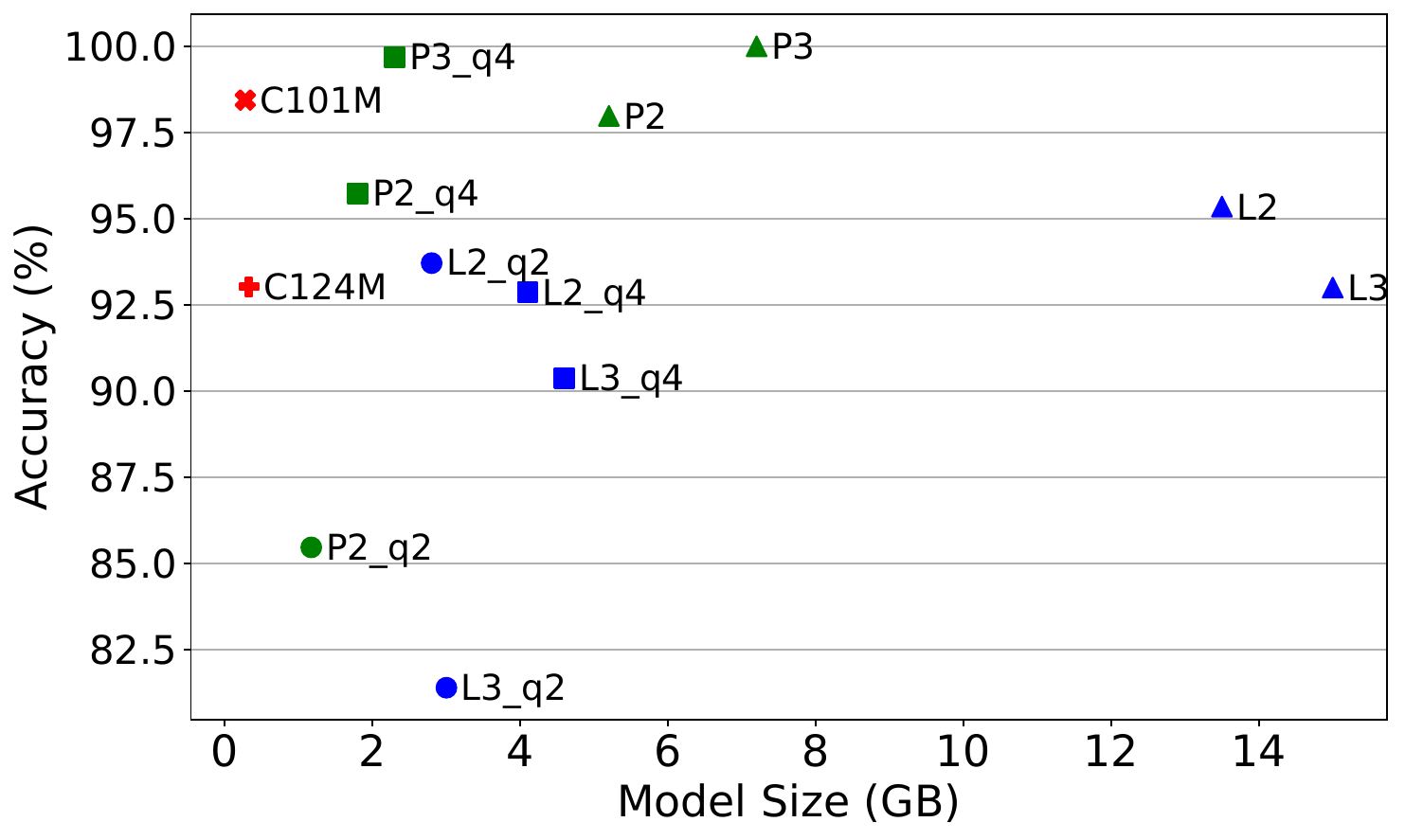}
      \vspace{-6mm}
        \caption{Gesture dataset}
     \vspace{-2mm}
        \label{fig:gesture_acc}
    \end{subfigure}
        \hfill
    \begin{subfigure}[b]{0.42\textwidth}
        \centering
  \includegraphics[width=\textwidth]{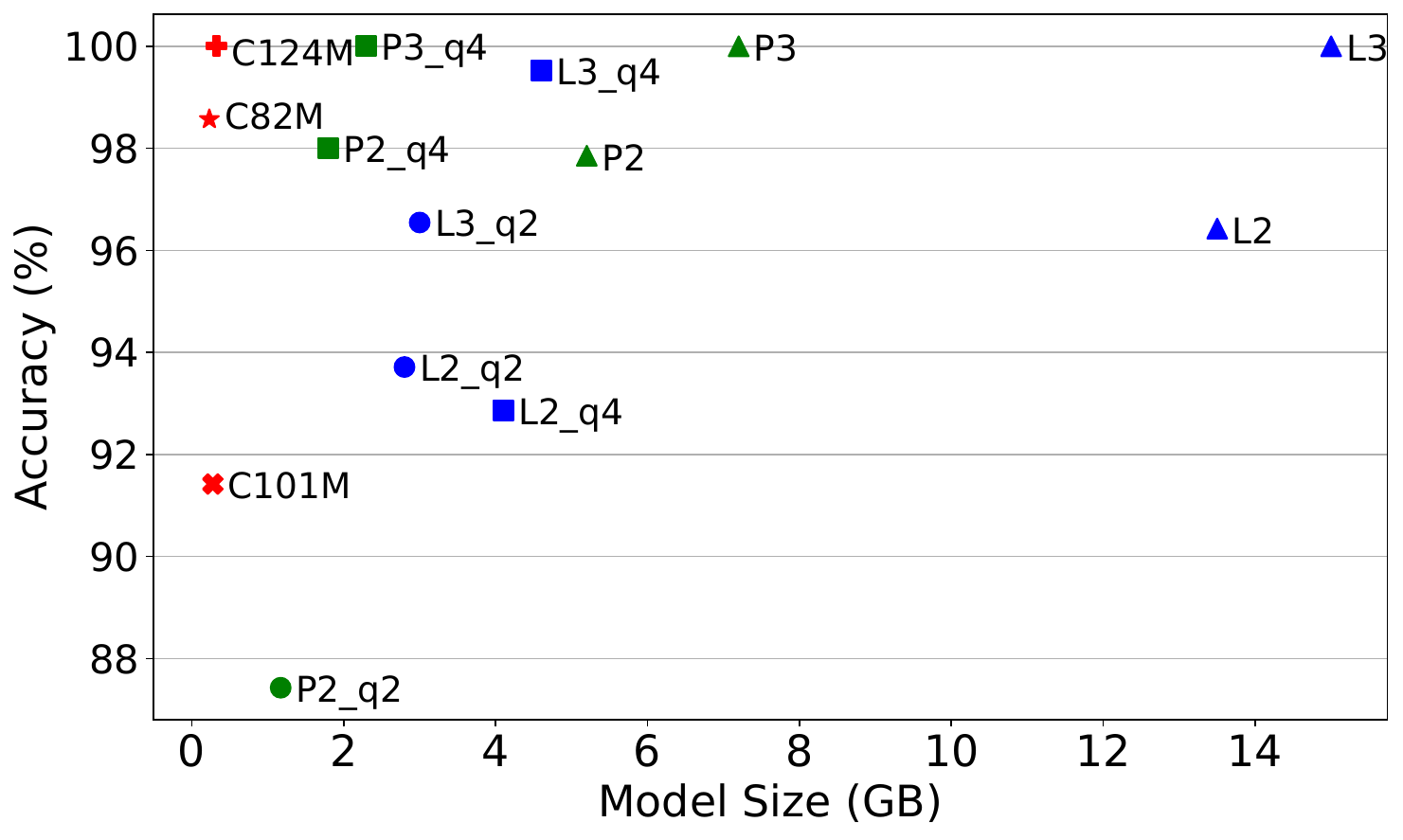}
      \vspace{-6mm}
        \caption{Localisation dataset}
     \vspace{-2mm}
        \label{fig:local_acc}
    \end{subfigure}
            \hfill
    \begin{subfigure}[b]{0.15\textwidth}
        \centering
  \includegraphics[width=\textwidth]{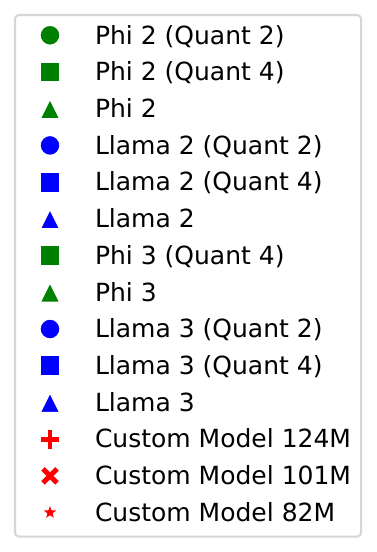}
      \vspace{+2mm}
        \label{fig:legend}
    \end{subfigure}

    \caption{\emph{Compares the accuracy of fine-tuned off-the-shelf models (Phi and Llama) and custom models across (a) gesture and (b) localisation datasets, as a function of model size. The plots show that larger models do not always achieve high accuracy. In several cases, a smaller custom model achieves comparable or better performance than the larger model}}
    \vspace{-2mm}
    \label{fig:accuracy}
    \end{figure*}

\fakepar{Experiment setup} We use multiple single-board computers for deploying custom and off-the-shelf models. This includes Latte Panda Sigma~\cite{lattepanda} (Intel Core i5-1340P, 32 GB RAM, Ubuntu 20.04.4 LTS), Orange Pi 5 \cite{orangepi5} (16GB RAM, Orange Pi OS), and Orange Pi Zero 2 W \cite{orangepi502w} (2GB RAM, Orange Pi OS). We use the llama.cpp~\cite{llama.cpp} library and its metrics (token generation rate and run time) for the execution and evaluation of the selected models. 

We have set the LLM's parameters for generating the responses: temperature (T = 0.7), repeat penalty = 1.1, and threads (t = 2) to maintain consistency across models and experiments. In addition to the custom models, we employ off-the-shelf LLMs and their quantized versions \cite{quantization} to enable comparison. In the quantized versions, model weights are stored at lower precisions, which reduces the memory requirements but can also impact the quality of the model’s responses. Several quantization schemes exist, among which we used q2\_k and q4\_k for some of the evaluations. In all experiments, we conduct ten trials for each configuration and plot the average value and the standard deviation unless specified otherwise. We also evaluate model performance using accuracy, which is the percentage of correctly generated outputs. Since our prompts follow a specific template, the correct label is expected within the first few tokens. We check for the expected word within the first 3–4 tokens (ignoring line breaks). If it is missing, or if gibberish or incorrect classes are predicted, the output is classified as incorrect. Accuracy is calculated as the ratio of correct predictions to the total number of test cases, then multiplied by 100 for a percentage score.

\subsection{Accuracy of Fine-tuned Custom Models}

The custom models, Llama and Phi, are fine-tuned on three different datasets, and their accuracy is evaluated based on test data. The quantized versions of the fine-tuned Phi and Llama models are also evaluated for accuracy. However, the same analysis could not be performed on the swimming dataset, as the fine-tuning process for Llama 3 and Phi 3 exceeded 72 hours (3 days) due to the dataset's higher count of samples compared to the collected datasets. It can be noted that Phi 3\_q2 is not plotted as it achieved zero accuracy, as it repeatedly generated irrelevant outputs (mostly repeating the prompt). 

\fakepar{Insights} As shown in Figure \ref{fig:accuracy} the accuracy of the custom models, Llama, and Phi models fine-tuned with the collected datasets. Notably, the smaller custom models perform on par with or better than the larger models. Among the datasets, higher accuracy is observed for the localization dataset, suggesting that the gesture dataset poses more challenges. For the gesture dataset, Llama 2 outperforms Llama 3, while Phi 3 consistently performs well across both datasets, with Phi 2 following closely. Phi models performed better than Llama models, which can be attributed to their pre-training data being more focused on computer programming (coding) datasets, enhancing their ability to handle sensor data as well. 

\subsection{Different Pre-training Datasets}

\begin{figure}[htb]
    \centering
            \begin{subfigure}[b]{0.49\columnwidth}
        \centering
        \includegraphics[width=\textwidth]{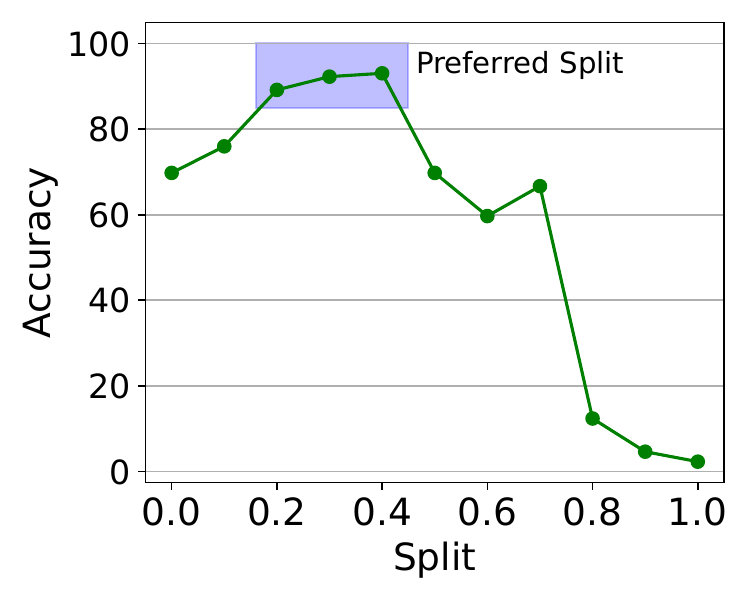}
        \vspace{-6mm}
        \caption{Gesture}
      \vspace{-3mm}
        \label{fig:split_gesture}
    \end{subfigure}
    \hfill
    \begin{subfigure}[b]{0.49\columnwidth}
        \centering
        \includegraphics[width=\textwidth]{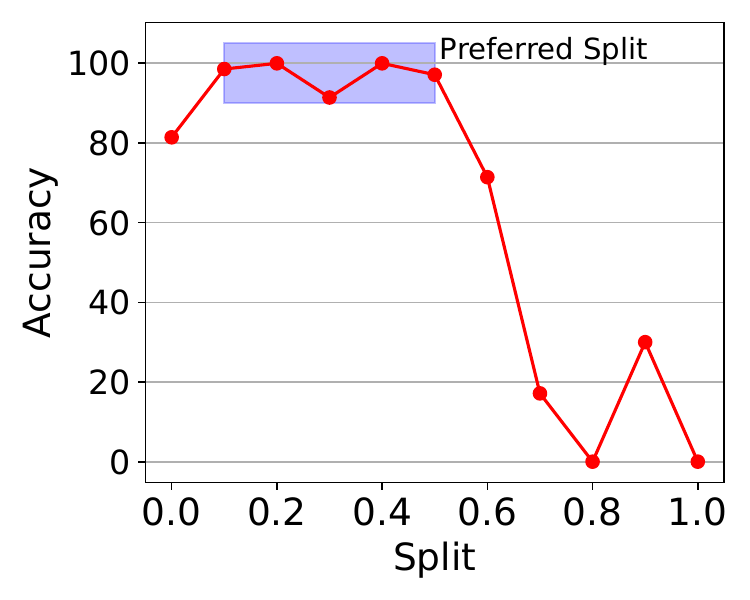}
      \vspace{-6mm}
        \caption{Localisation}
     \vspace{-3mm}
        \label{fig:split_local}
    \end{subfigure}

    \caption{\emph{Compares the accuracy of fine-tuned custom models (124M) on the (a) gesture and (b) localization datasets. The models are pre-trained on varying splits of sensor data and general web data, with a split of 0 indicating training solely on web data and a split of 1 indicating training exclusively on the sensor dataset. The shaded region highlights the preferred data split.}}

    \vspace{-2mm}
    \label{fig:splits}
    \end{figure}

 With the sensor dataset created earlier, we merged the Fineweb dataset to generate a series of mixed datasets in varying proportions, ranging from 0 (only Fineweb dataset) to 1 (only sensor dataset) in increments of 0.1. Then, a custom 124M model is pre-trained on these datasets separately and subsequently fine-tuned separately on gesture and localization datasets.Figure~\ref{fig:split_gesture} and \ref{fig:split_local} show the accuracy of the fine-tuned 124M parameter model, pre-trained on varying splits of sensor and web data. It can be observed that an almost equal mix of web and sensor data yields the highest accuracy (marked in blue). For both tasks, performance declines significantly as the data split shifts towards either extreme (i.e., pure web or pure sensor data).
 
 \fakepar{Insights} This suggests that a balanced dataset improves model performance for sensor data-specific applications like gesture recognition and localization. A notable spike in accuracy is observed with the localization dataset when the split is 0.9, which occurred because the model consistently returned the same output label for all input prompts, artificially inflating the accuracy to 33\%.

\subsection{Varied Custom Model Parameter}

\begin{figure}[h]
\centering
  \includegraphics[width=0.8\columnwidth]{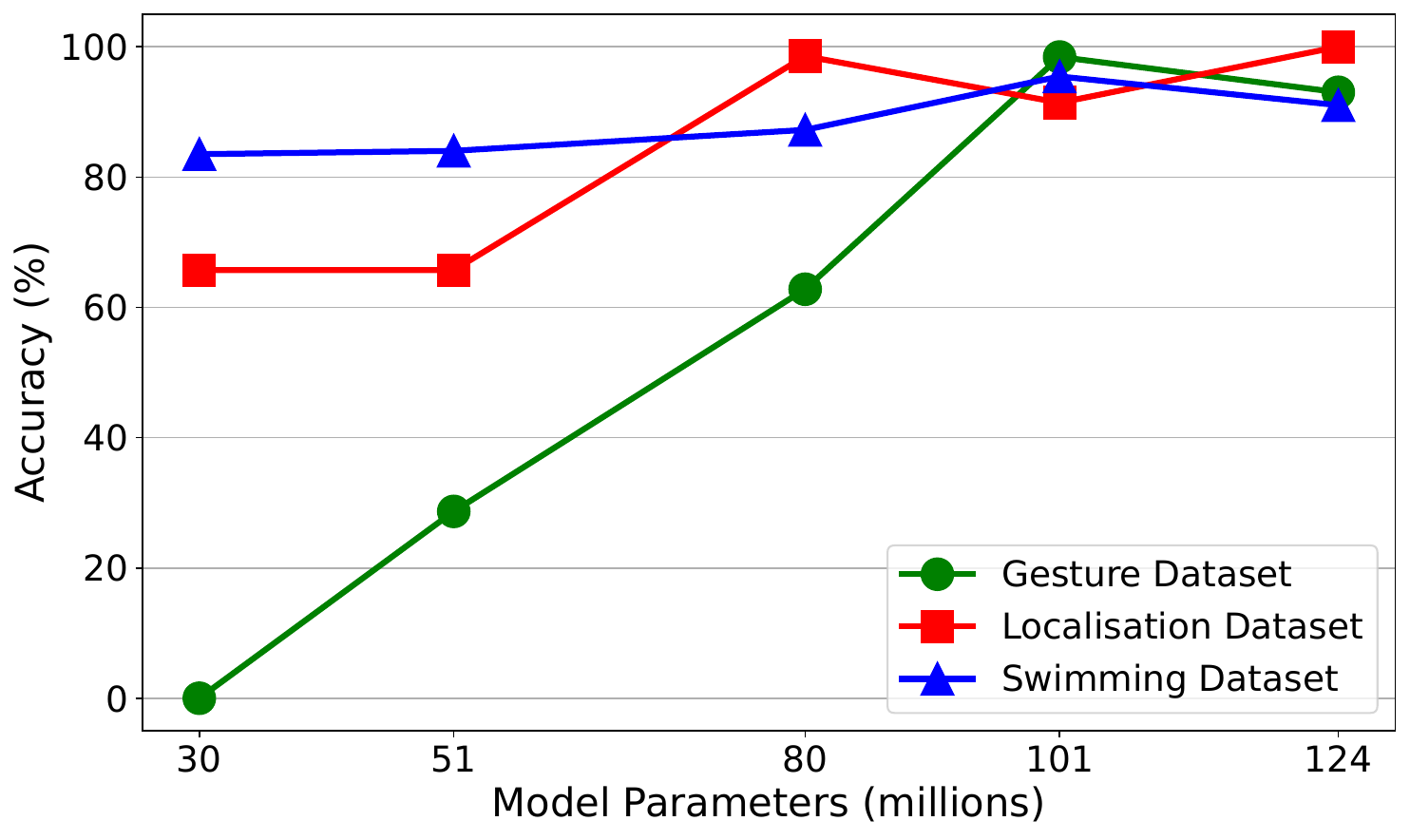}
  \vspace{-4mm}
  \caption{\emph{The accuracy of sensor data analysis increases with the model’s parameter size. Notably, even smaller models with fewer than 100 million parameters achieve high accuracy.}}
  \vspace{-4mm}
  \label{fig:gpt_acc}
\end{figure}

\begin{figure}[htb]
\centering
  \includegraphics[width=0.9\columnwidth]{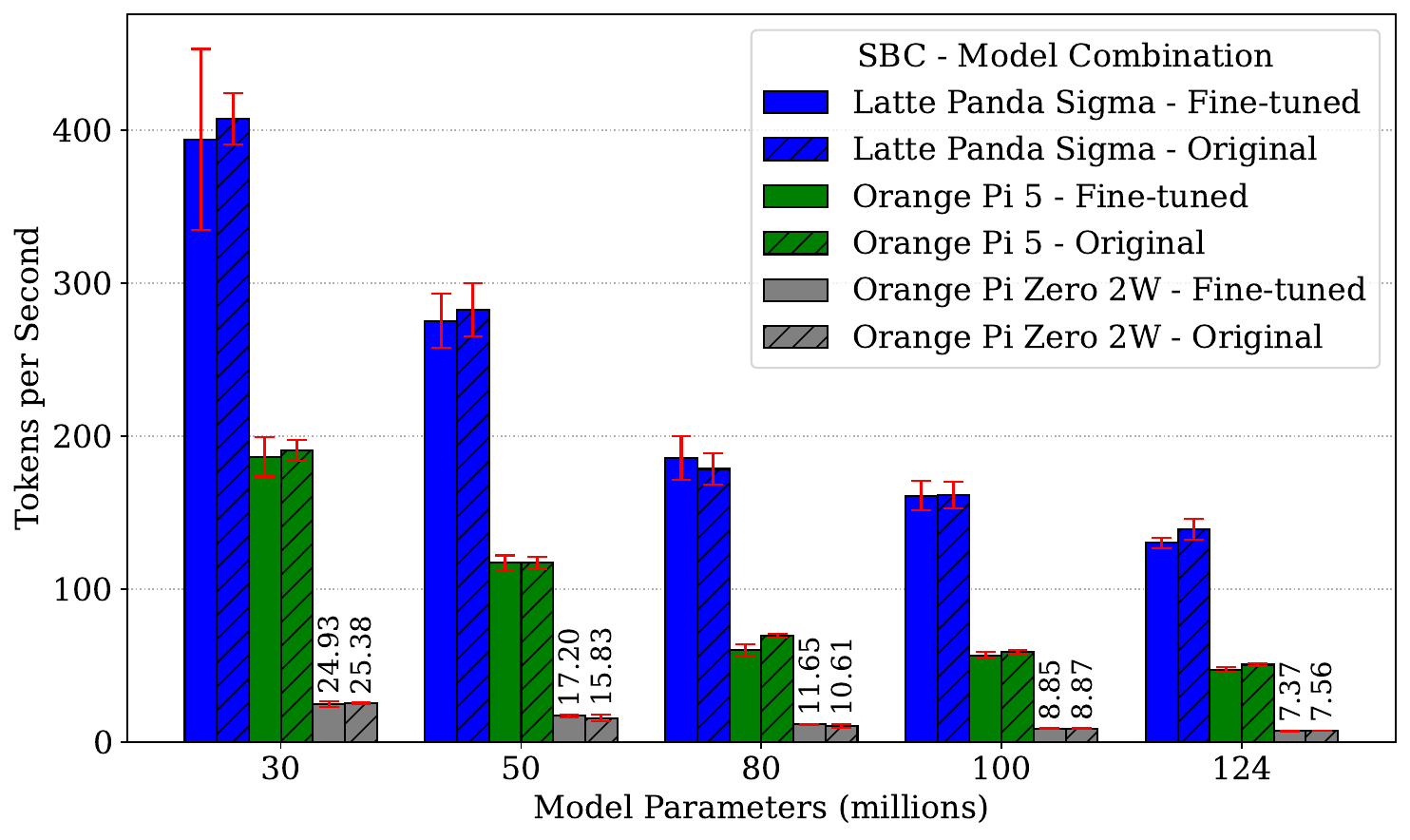}
  \vspace{-4mm}
  \caption{\emph{Shows the variation of evaluation tokens per second when prompted with a gesture recognition prompt.  The lowest token generation rate is comparable to the average human typing speed, demonstrating that custom models achieve reasonable performance, even on resource-constrained platforms.}}
    \vspace{-4mm}
  \label{fig:small_models}
\end{figure}

\begin{figure}[htb]
\centering
  \includegraphics[width=0.9\columnwidth]{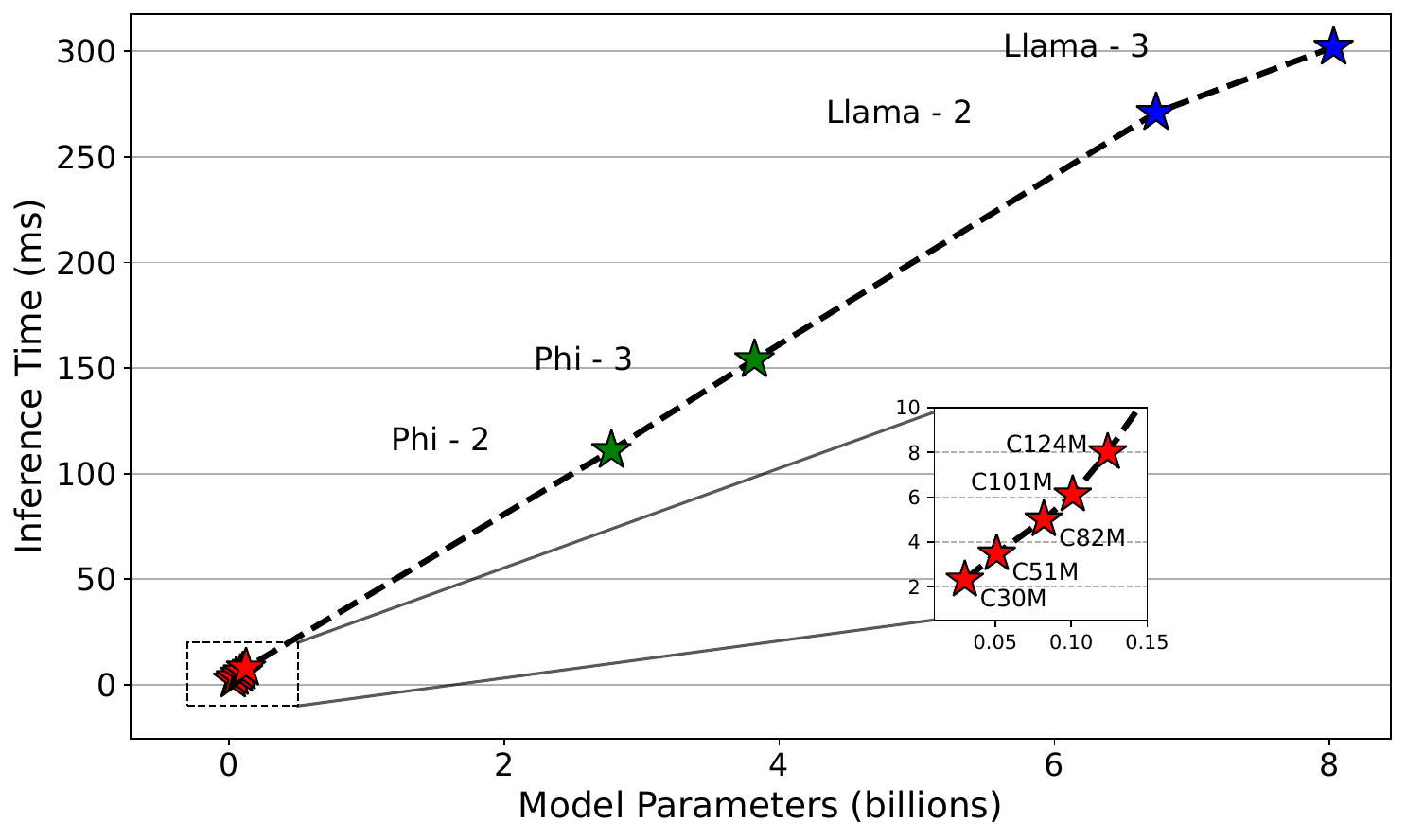}
  \vspace{-4mm}
  \caption{\emph{Smaller models enable rapid inference. \system\space-trained models significantly improve inference time while maintaining high accuracy for the sensor data analysis.}}
  \vspace{-4mm}
  \label{fig:inference_timings}
  \Description{}
\end{figure}

In evaluating the performance of custom models on resource-constrained platforms, we employ models with varying parameters (30M to 124M), fine-tuned on the datasets discussed above. We evaluate performance based on accuracy, inference time, and tokens per second. Figure~\ref{fig:gpt_acc} shows the accuracy of smaller models across the three datasets considered. We observe that accuracy generally improves as the number of model parameters increases, although some minor exceptions exist. Notably, the 30M and 50M models perform poorly on the gesture dataset, with the custom 30M model achieving zero accuracy. Figure~\ref{fig:small_models} shows the variation of token generation rate with the number of parameters of gesture fine-tuned custom models. We observe that models with fewer parameters achieve higher tokens per second rates compared to larger models or those with higher parameter counts. Additionally, the token generation rate decreases as the computational power of the device decreases. Figure~\ref{fig:inference_timings} shows the inference time with varied model parameter sizes across custom and off-the-shelf models.

\fakepar{Insights} Smaller models achieve higher token generation rates and reduced inference times, presenting a trade-off with accuracy.

As shown in Figure~\ref{fig:gpt_acc} for the swimming dataset, while the maximum accuracy achieved was 93.1\%, the highest F1-score recorded was 0.78, which is lower than the 0.97 F1-score reported in \cite{swimming_dataset}. It is important to note that we only used 3 datastreams (accelerometer readings along X, Y, and Z) out of the 11 provided in the dataset, due to the limited window context of the smaller models used in this study.


\subsection{Multiple Active Instances}

\begin{figure}[htb]
\centering
  \includegraphics[width=0.9\columnwidth]{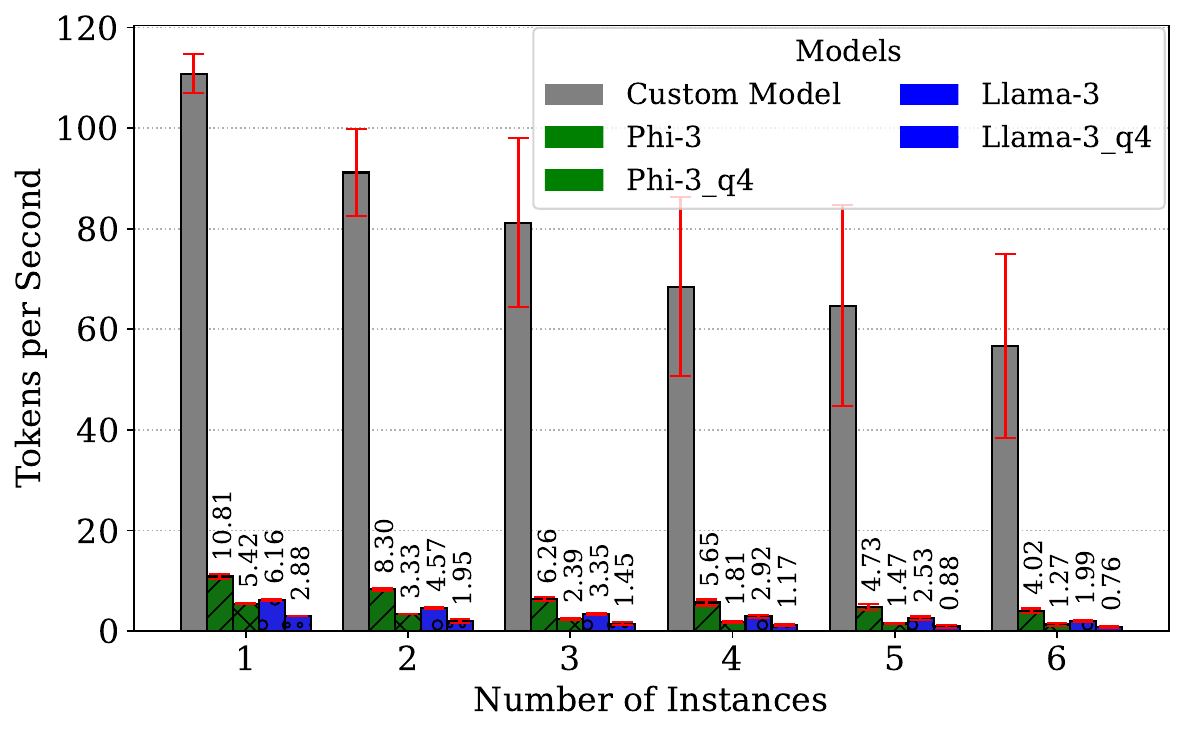}
  \vspace{-4mm}
  \caption{\emph{Shows the variation of evaluation tokens per second when multiple instances of the same LLM are running simultaneously on LattePanda Sigma.}}
    \vspace{-4mm}
  \label{fig:multiple_active_instances}
\end{figure}

We evaluate the impact of running multiple concurrent instances of the same model on the token generation rate. We use the sample prompt: \textit{"How to interface a sensor to a micro-controller? Explain in great detail."} and set the number of new tokens to generate to $n = 300$, deploying the models on LattePanda Sigma.As shown in Figure \ref{fig:multiple_active_instances}, we observed that across all models, the overall token generation rate decreases as the number of active instances increases. This rate reduction is more pronounced in larger models compared to smaller ones. 

\fakepar{Insights} Notably, the custom model exhibits a significantly higher token generation rate due to its smaller size, which may allow multiple custom models specialized for different sensor data-related tasks to operate concurrently on resource-constrained devices without significantly compromising performance.

\subsection{Varied Background Load}

\begin{figure}[htb]
\centering
  \includegraphics[width=0.9\columnwidth]{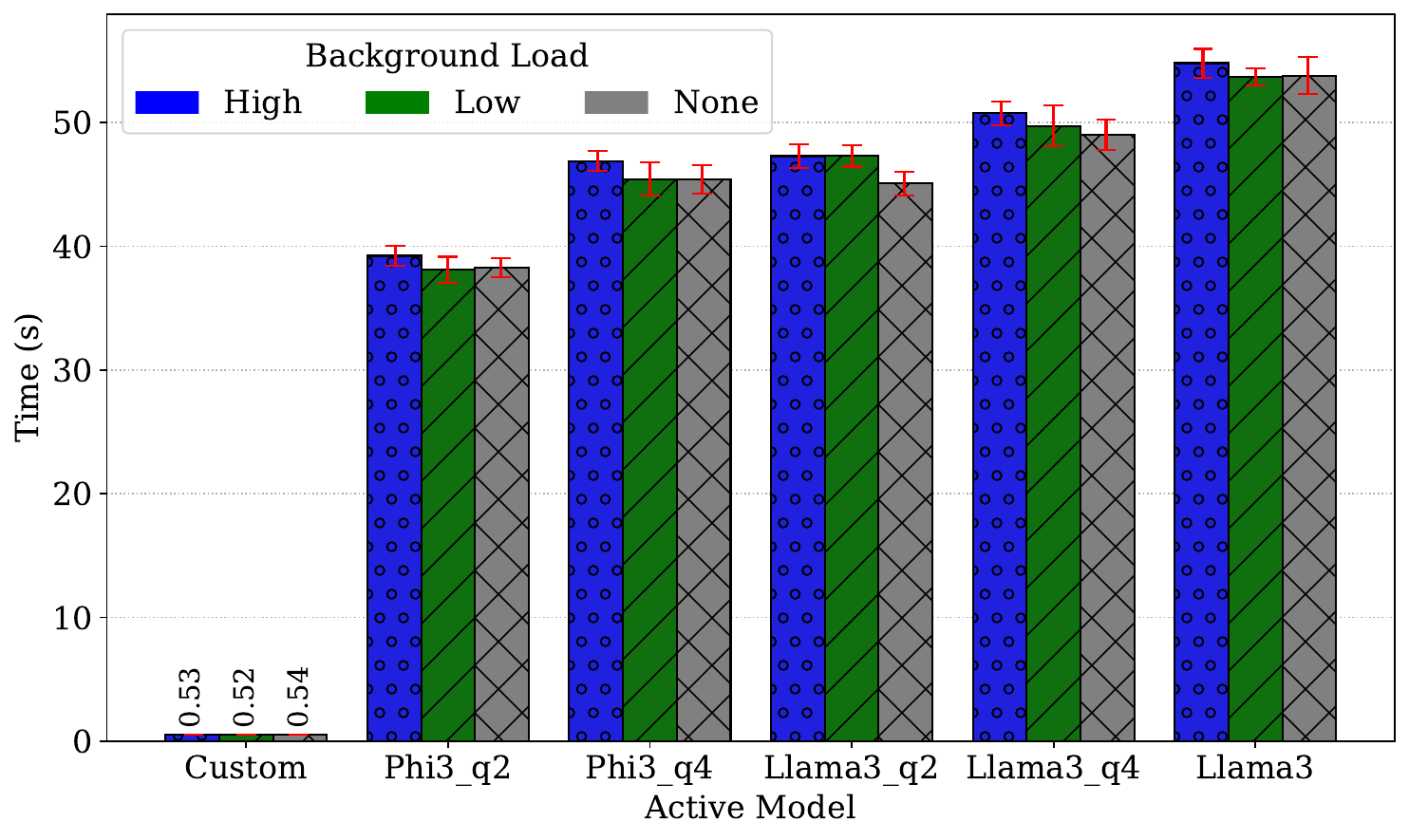}
  \vspace{-4mm}
  \caption{\emph{Displays the total time taken by different models for inferring location under various background load conditions on the LattePanda Sigma. Custom models significantly outperform the other models by completing the inference task within a second.}}
    \vspace{-4mm}
  \label{fig:background_load}
\end{figure}

We evaluate the impact of background load on task completion time. To do this, we actively prompt a single model while multiple models are loaded into the memory of the LattePanda Sigma. We employ custom and off-the-shelf models fine-tuned on the localisation dataset, using a localisation-based prompt, and set the number of new tokens to generate to $n = 9$. We consider three different background loads based on the inactive models: High (three instances of Llama-3), Low (three instances of Phi-3), and None (no inactive LLMs loaded).As shown in Figure~\ref{fig:background_load}, we observe that the background load from inactive LLM instances has little to no impact on the time taken to achieve a task-inferring location. 
\fakepar{Insights} The time taken by the custom model to perform the task is significantly less (over 70 times) than the time taken by other models for the same task across all background load conditions considered. This complements the results of the previous experiment, which showed that the custom model had a significantly higher token generation rate.

\subsection{Multiple Edge platforms}

\begin{figure}[htb]
\centering
  \includegraphics[width=0.9\columnwidth]{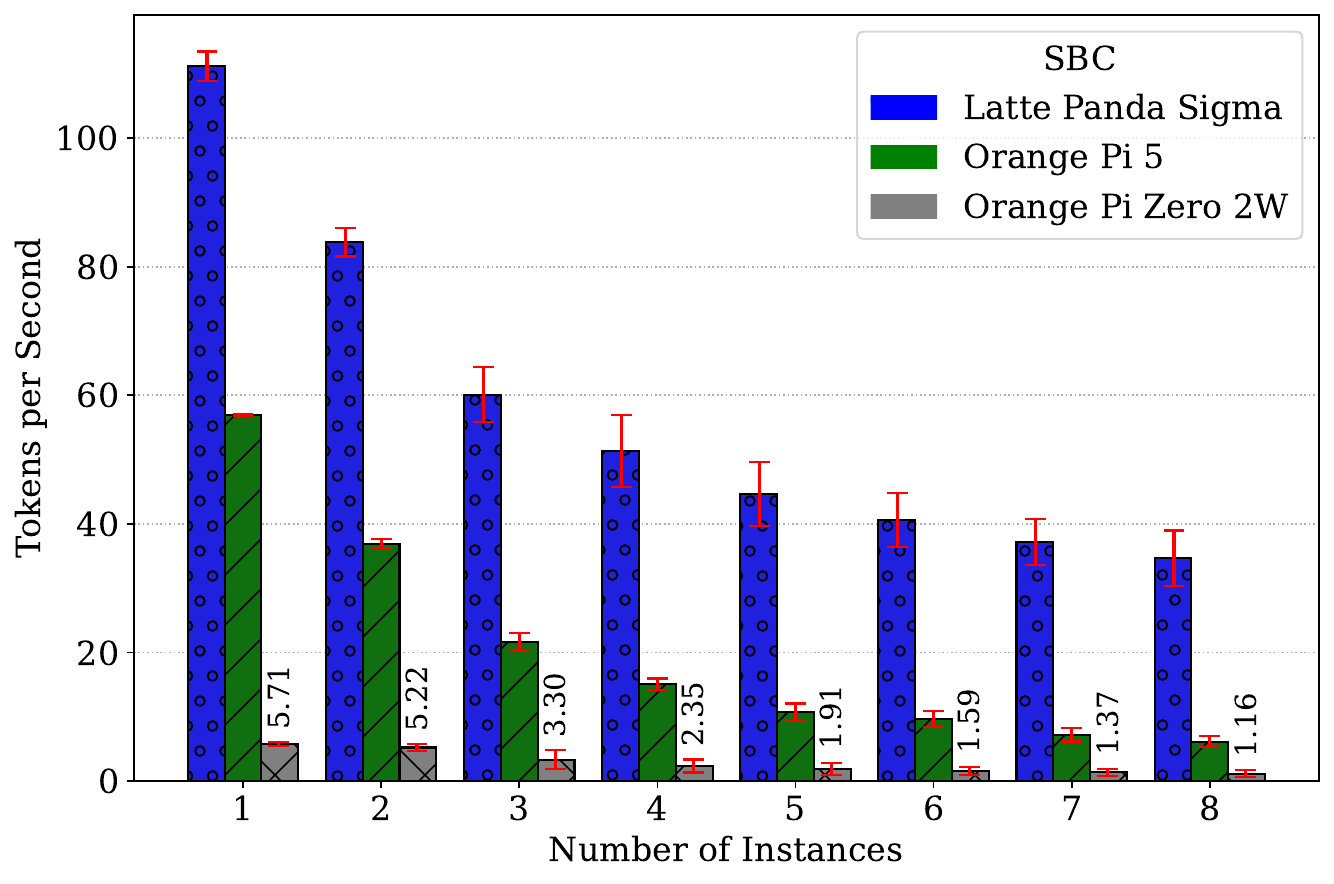}
  \vspace{-4mm}
  \caption{\emph{The smaller size of these models enables concurrent loading of multiple specialized models. Their token generation rates show they maintain efficient inference speeds even when running multiple instances simultaneously.}}
    \vspace{-4mm}
  \label{fig:tokens_load}
\end{figure}

We evaluated the performance of the custom model on different single-board computers and assessed it based on average tokens per second and the impact of running multiple concurrent instances of the same model. We used a custom model fine-tuned on the localisation dataset, with a localisation based prompt and a new tokens generated parameter set to $n = 4$. As shown in Figure~\ref{fig:tokens_load}, we observe that while the token generation rate decreases with an increasing number of concurrent instances on the SBC, it maintains a reasonable rate. Although the custom model can be deployed on all the SBCs considered, we observe a steep drop in the token generation rate when moving to more resource-constrained devices. This drop is due to the decreasing computing power of the processors in the SBCs.

\fakepar{Insights} Even in highly constrained devices such as the Orange Pi Zero 2W, on which the considered off-the-shelf models cannot be deployed, the custom model achieves a token rate of approximately 6 tokens per second, comparable to the average typing speed of humans.

\section{Limitations and Discussion}

\fakepar{Model Architectures} Currently, the framework supports only pre-training of GPT-2 based architectures, although fine-tuning supports many other models.

\fakepar{Context Window length} The input prompt size is currently limited to 1024, which might limit the applicability of some usecases consisting of longer prompts

Here, the smaller models (below 120M) are scaled down from the 124M parameter model by reducing the number of transformer blocks but maintaining the relation $C = 64l$ as shown in Table~\ref{tab:gpt2_parameters}. However, it might be interesting to explore the training of smaller models created by varying $C$ and $l$ without constraining.

\section{Related Work}

\fakepar{Sensing and Models}
While it is widely known that LLMs excel at language-based tasks, various attempts are made to test LLMs on different modalities like images, audio, and time-series data. There are significant advancements in audio and image domains~\cite{audio_transformer, vision_transformer}. Time-series as input to LLMs still remains a bigger challenge to be solved, although there are many works aimed that have had decent progress~\cite{times-series-llm}. Advances in this benefit many domains like medical~\cite{health_learners} and sensor data analysis~\cite{penetrative_ai}, which primarily contain time-series data from sensors. Mo et al.~\cite{iot-lm} makes LLMs comprehend sensory data by modifying the LLM's architecture. A new multisensory multi-task adapter layer is introduced, making the model capable of perceiving eight IoT tasks. 

\fakepar{LLMs and Programming} Recent years have seen growing interest in using LLMs in the software development process. They demonstrate an increasing ability to generate relevant code from natural language prompts. LLMs are also used for other coding tasks like completion, syntax correction, and refactoring. These capabilities have led to surprising results: AlphaDev~\cite{AlphaDev2023}, for instance, discovered a faster sorting algorithm that surpasses previously known human benchmarks. Meta's LLM Compiler~\cite{MetaCompiler2024}, designed for compiler optimization, is another breakthrough by enhancing code generation efficiency and aims to optimize code for better performance and resource utilization. Consequently, alongside larger, cloud-based LLMs such as ChatGPT and Claude, smaller LLMs designed specifically for coding tasks have also emerged.
Examples include CodeLlama~\cite{roziere2023code}, StarCoder~\cite{llm_starcoder}, Codestral~\cite{llm_codestral}, DeepSeek Coder~\cite{llm_deepseek}, and CodeBERT~\cite{feng2020codebert}. Many of these LLMs are now part of commercial products, including GitHub CoPilot\footnote{\url{https://github.com/features/copilot}} and OpenAI Codex\footnote{\url{https://openai.com/index/openai-codex/}}. \system\space is complementary to these systems and can utilize LLMs optimized for coding purposes. 
\section{Conclusion}
\system\space enables the pre-training and fine-tuning small language models on custom user data. The framework supports deploying models at the edge, ranging between 30M and 124M parameters. Our results show that these smaller models can match or even surpass the performance of much larger models across various applications. Incorporating domain-specific pre-training data further enhances their effectiveness. This framework takes a step towards deploying smaller, domain-adapted language models optimized for edge computing to support embedded sensing applications.

\section{Acknowledgements}
This work is funded by a grant from  NUS-NCS Joint Center~(A-0008542-02-00). It is also supported by a startup grant (A-8000277-00-00), a MoE Tier 1 grant (A-8001661-00-00) and an unrestricted gift from Google through their Research Scholar Program~(A-8002307-00-00),  hosted at NUS.

\bibliographystyle{plain}
\FloatBarrier
\bibliography{ref}

\end{document}